\pdfoutput=1
\documentclass[11pt]{article}

\usepackage[final]{acl}
\usepackage{times}
\usepackage{latexsym}
\usepackage{graphicx} 
\usepackage{booktabs}
\usepackage[T1]{fontenc}
\usepackage[utf8]{inputenc}
\usepackage{amsmath}
\usepackage{microtype}
\usepackage{verbatim}
\usepackage{inconsolata}
\usepackage{booktabs, multirow}

\title{Reverse Probing: Supervised Token-level Uncertainty Quantification for Large Language Models in Clinical Text}
\author{Bushi Xiao \\
  University of Florida \\
  Gainesville, FL\\
  \texttt{xiaobushi@ufl.edu} \\\And
  Sarvesh Soni \\
  U.S. National Library of Medicine \\
  Bethesda, MD \\
  \texttt{sarvesh.soni@nih.gov} \\\And
  Daisy Zhe Wang \\
  University of Florida \\
  Gainesville, FL \\
  \texttt{daisyw@cise.ufl.edu}
  }

\begin{document}
%\linenumbers
\maketitle
\begin{abstract}
As large language models are increasingly deployed for clinical text, ensuring they can reliably signal their own uncertainty becomes critical. Most existing uncertainty quantification (UQ) methods are designed for open-domain generation and cannot localize uncertainty at the token or span level in long clinical text. We propose Reverse Probing, the first uncertainty quantification framework specialized for clinical summarization, which estimates token-level uncertainty directly from pre-existing clinical summaries. Rather than sampling new outputs, Reverse Probing treats the text as a probe into the model's internal state, extracting uncertainty features from four categories of signals. We evaluate Reverse Probing on two expert-annotated clinical datasets and outperform eight adapted baselines, achieving up to $4\times$ higher AUPRC while running over $15\times$ faster than sampling-based methods. Feature analysis reveals that the log-logit difference induced by the source record and the neighborhood context are the most consistent predictors across all models. This study offers interpretable insights into how models internally respond to unsupported clinical content.
 
\end{abstract}

\section{Introduction}

Uncertainty quantification (UQ) for Large Language Models (LLMs) has been widely studied in general domains, yet its application to clinical text remains limited. Clinical text presents distinct challenges from general tasks: it is dense with terminology, errors often occur at a single word or number within an otherwise correct sentence, and small mistakes can carry serious consequences. Most existing UQ methods operate at the case level and are poorly suited for fine-grained, token-level uncertainty in long clinical text.

UQ is also fundamentally distinct from hallucination detection. Rather than using external methods to detect whether a model's output is factually correct, UQ asks how confident the model is in its own predictions. This is a form of self-assessment that reflects model reliability independent of ground truth. \citet{yona2026} formalizes this distinction, redefining hallucination as a confident error and arguing that models should express honest uncertainty instead of optimizing solely for accuracy. 

Consequently, token-level UQ offers a more actionable pathway for model alignment and refinement. The prevailing assumption is that UQ requires generation: sample the model's outputs multiple times and treat divergence as a proxy for uncertainty. This method works for short question-answering tasks, but is unsuitable for clinical settings, where generating new clinical notes is hard to evaluate reliably, repeated sampling over long documents is computationally expensive, and expert-annotated clinical text is scarce.

Traditionally, token-level probing is designed for encoder-only transformer models like BERT \citep{devlin-etal-2019-bert}, which evaluate confidence by predicting a masked token within a bidirectional context. Decoder-only LLMs operate auto-regressively and do not natively support such token-masking method. We bridge this architectural gap with a teacher forcing strategy to imitate the mask prediction on bi-direction models. Though the method is not equivalent to a native MLM prediction, the model is mimicking a pseudo mask prediction task. 

Reverse Probing reflects a deliberate inversion of the standard UQ paradigm. Conventional methods ask the model to generate text and then evaluate the output. We reverse this direction: instead of examining what the model produces, we feed existing clinical text back into the LLM to extract a comprehensive snapshot of its internal activations. A well-supported token inherently aligns with and anchors to the relevant clinical records within the representation space. Conversely, when a token is unsupported, this factual grounding is absent. We harness this absence of alignment as a latent uncertainty signal, effectively turning the text itself into a probe of the model rather than a product of it.

Our contributions are as follows:
\begin{itemize}
    \item We introduce Reverse Probing\footnote{\url{https://github.com/kitayamachingtak/ReverseProbing}}, the first token-level UQ framework for clinical summaries. Our framework eliminates multi-sampling to reach an exceptional computational efficiency.
    \item We establish the first benchmark for UQ on clinical summaries. We conduct a systematic evaluation comparing Reverse Probing against eight baselines across six LLMs on two expert-annotated datasets.
    \item We analyse where the uncertainty signal comes from. The log-logit
    difference induced by the source record (\texttt{delta\_energy}) and local
    neighborhood context are the most consistent predictors across models and
    feature configurations.
\end{itemize}

\begin{figure*}[t]
    \centering
    \includegraphics[width=\linewidth]{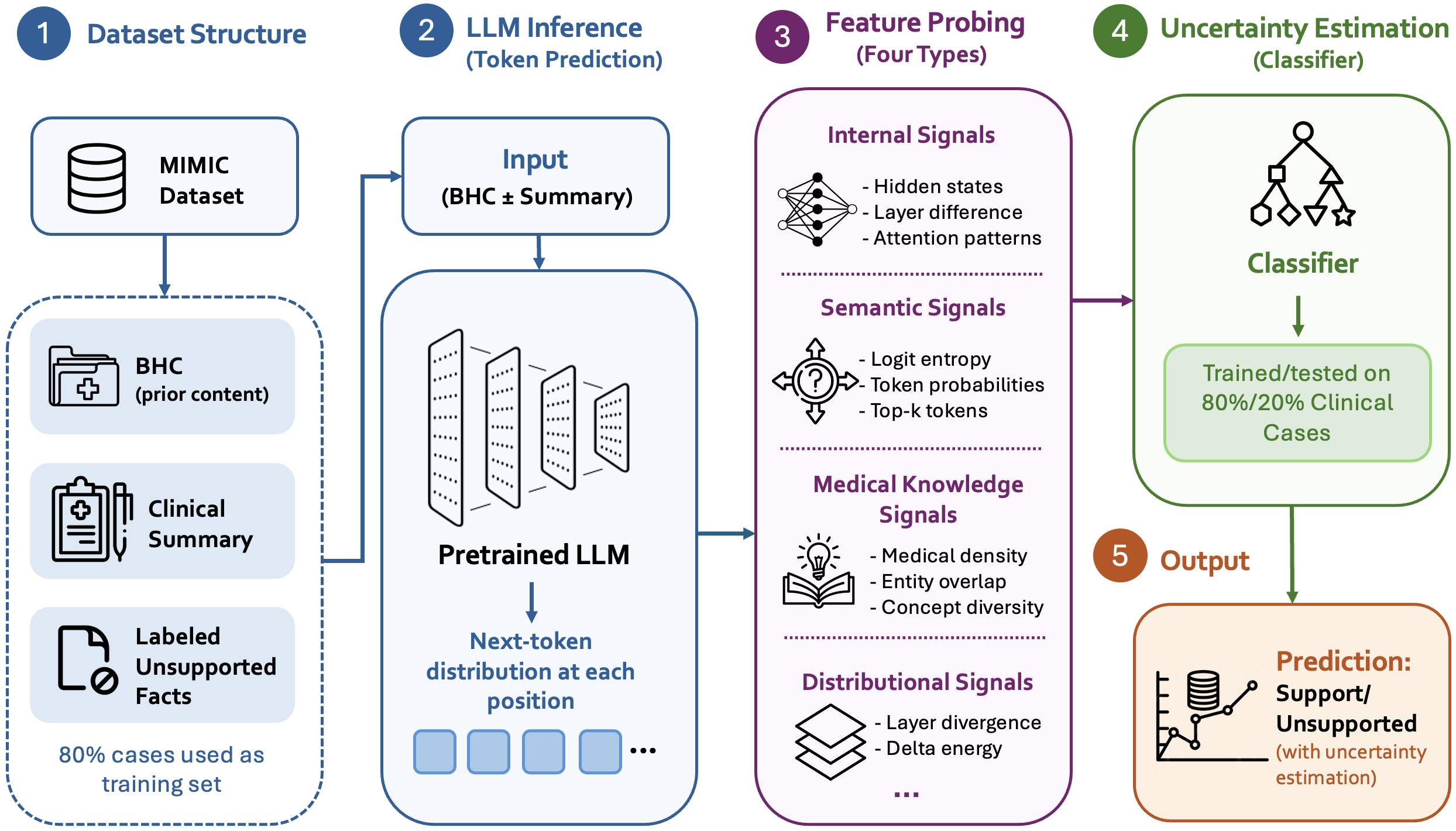}
    \caption{Overview of the Reverse Probing framework. Each summary is passed through a frozen LLM twice, once with the BHC text and once without, yielding distributional, semantic, internal, and medical-knowledge features at every token position. A supervised classifier then predicts token-level uncertainty.}
    \label{fig:framework}
\end{figure*}

\section{Related Work}
\subsection{Uncertainty Quantification}

Traditionally, uncertainty quantification relies on sampling-based approaches. These methods require the model to generate multiple output candidates. The uncertainty is then quantified from the variance or distribution of the multiple generated text. SelfCheckGPT \citep{manakul2023selfcheckgpt}, Semantic Uncertainty \citep{kuhn2023semantic}, and Semantic Entropy \citep{farquhar2024detecting} follow this paradigm. \citet{hou2024decomposing} further decomposes uncertainty into aleatoric and epistemic components. However, these methods require task-level new text generation without localizing uncertainty to specific tokens or spans.

Several recent methods derive uncertainty from model internals. \citet{liu2024uncertaintyestimationquantificationllms, vazhentsev2025uncertainty, li2025uqac} show that hidden activations and attention heads encode uncertainty information. EUQ \citep{huang2026detecting} on evidential uncertainty also performs on internal signals, which decomposes uncertainty into conflict and ignorance to flag misbehaving vision-language model responses at the level of a full generation. Among all methods above, only Semantic Entropy (SE) Probes \citep{kossen2024sep} adopts a supervised formulation, which trains a lightweight probe on hidden states to predict the semantic entropy.

Despite these advances, all existing methods are designed and evaluated on general-domain tasks, and score a generation as a whole. Unsupported content in clinical summary is sparse and crucial, with specialized terminologies, so uncertainty has to be localized rather than aggregated into a paragraph-level single score. More importantly, the summaries are pre-existing text rather than being generated by the model. No existing method fits this setting, which requires working post hoc on a fixed summary, localizing at the span level, and operating without access to the generation process.

\subsection{Clinical Explainability and Faithfulness}
Explainability methods have been widely applied to clinical decision support, but rarely to clinical uncertainty itself. SHAP \citep{lundberg2017unified} remains the standard framework for feature-level explanation, while \citet{hur2025comparison} shows that natural language explanations outperform raw SHAP scores for clinical decision-making. \citet{qureshi2025explainability} find that no single XAI method dominates across all clinical criteria. Several works argue that combining UQ with explainability improves reliability in high-stakes settings \citep{salvi2025explainability, dubey2025ubiqtree, lu2024uncertainty}. However,  none of the work applies this combination specifically in clinical text.

Faithfulness evaluation for clinical summaries also presents a related challenge. Standard metrics correlate poorly with expert judgment \citep{adams2023meta}, and DocLens \citep{xie2024doclens} improves this with claim-level evaluation. \citet{croxford2025evaluating} validate LLM-as-a-Judge as a scalable alternative to human annotation. \citet{wang2025semantic} uses semantic consistency across sampled outputs as a semantic signal for radiology reports, but their approach still depends on sampling and operates at the sentence level. Token-level faithfulness evaluation over clinical text remains an open problem.

\section{Methodology}
In clinical summaries, the majority of the sentences are factually supported, while unsupported content appears only in sparse, scattered spans. This makes fine-grained UQ fundamentally different from case-level evaluations, where standard methods are too coarse to capture localized errors.

Our approach is inspired by masked probing for bidirectional encoder models, where a token is hidden and the model is asked to recover it from its surroundings. That estimate is exactly what we want. However, decoder-only LLMs are auto-regressive and offer no masking mechanism. Teacher forcing gives us the same quantity without one: at every position the model predicts the token rather than reading it, so each token is effectively masked from its own prediction. This pseudo-masking turns a generative model into a token-level probe without modifying its weights.

The key insight behind Reverse Probing is to treat the source record as an attention anchor. In clinical domain, each discharge summary is written from the Brief Hospital Course (BHC), a long clinical note documenting the patient's admission and treatment. When a summary token is well supported, the model's attention concentrates on the BHC and its prediction sharpens once the record is present. When a token is unsupported, that concentration is absent or misdirected, and removing the BHC changes little. We therefore read the absence of grounding as a signal in its own right, by comparing activations with and without the BHC. This is precisely where the "reverse" in Reverse Probing comes from. While traditional methods generate new text to check for factual accuracy, our approach does the opposite: it uses the existing text as a probe to locate the key features and uncover exactly where the model's internal layers are struggling to align with clinical evidence.

Concretely, we feed the clinical summary with and without the BHC into a frozen LLM as shown in Figure~\ref{fig:framework}, extract features from its internal activations and output distributions, and train a classifier to predict which part of the internal representation will be activated when encountering an unsupported span. We organize features into four categories described in Section~\ref{sec:features}.

\subsection{Dataset}

\begin{figure}[t]
    \centering
    \includegraphics[width=\linewidth]{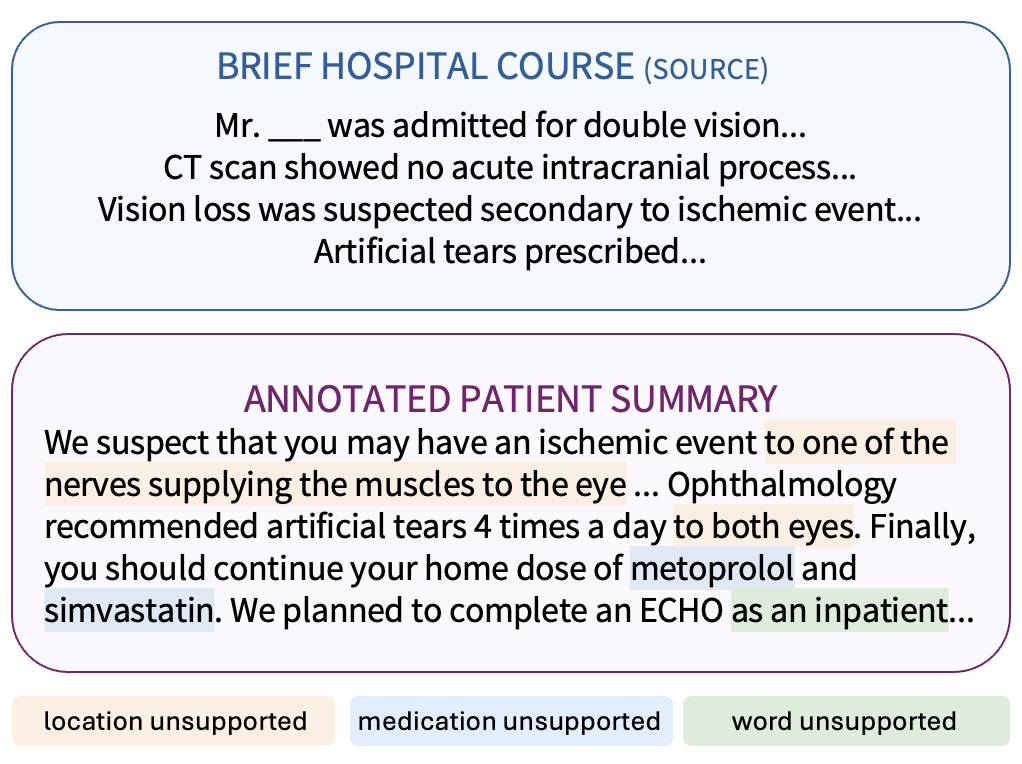}
    \caption{An example from Hallucinations-MIMIC-DI. The BHC serves as the source for the patient discharge summary. Highlighted spans in the summary mark unsupported facts annotated by experts, colored by label types. Dataset Hallucinations-Generated-DI follows the same structure.}
    \label{fig:dataset}
\end{figure}

This work uses Hallucinations-MIMIC-DI and Hallucinations-Generated-DI introduced by \citet{Hegselmann}. Both datasets are derived from MIMIC-IV-Note \citep{PhysioNet-mimic-iv-note-2.2}, a large-scale corpus of de-identified clinical notes from the Beth Israel Deaconess Medical Center. Each clinical case consists of a BHC paired with a discharge summary. The BHC serves as the source document recording the patient's clinical course, while the summary condenses this information for the patient. Hallucinations-MIMIC-DI contains 100 doctor-written discharge summaries, and Hallucinations-Generated-DI contains 100 LLM-generated summaries \citep{hegselmann2024data}, while the BHC text from both datasets originates from MIMIC-IV-Note. Both datasets provide span-level annotations of unsupported facts by two trained medical experts, where each annotation records the start and end offsets of the unsupported span with error types, as illustrated in Figure~\ref{fig:dataset}. These two datasets are highly imbalanced. In Hallucinations-MIMIC-DI, unsupported spans account for 7.33\% of total tokens with an average of 2.86 spans per summary, whereas in Hallucinations-Generated-DI the ratio drops to 2.05\% with an average of 1.14 spans per summary.

\subsection{Feature Extraction}
\label{sec:features}

We feed the existing summary back through the model under teacher forcing, as shown below.

\vspace{10pt}
\fbox{\parbox{0.92\columnwidth}{\footnotesize
\ttfamily Brief Hospital Course: \{BHC\}\\[4pt]
\rmfamily
\begin{tabular}{@{}r@{\ }l@{\ }c@{\ }l@{}}
$i=1$: & \texttt{Summary:} & $\rightarrow$ & $\mathbf{p}_1$\\
$i=2$: & \texttt{Summary: The} & $\rightarrow$ & $\mathbf{p}_2$\\
$i=3$: & \texttt{Summary: The patient} & $\rightarrow$ & $\mathbf{p}_3$\\
\multicolumn{4}{c}{$\vdots$}\\
\end{tabular}
}}
\vspace{5pt}

At position $i$ the model gives a distribution over the vocabulary,
$\mathbf{p}_i = P(\cdot \mid \text{BHC}, t_{<i})$, conditioned on the clinical
record and on the summary tokens that came before. We read that distribution
at every position, along with the hidden states and attention rows there, and
turn each position into one feature vector. Causal masking means position $i$
depends only on its own prefix, so a single pass covers the whole summary at
once. Reverse Probing thus works at the token level while costing two forward
passes per summary, no matter how long the summary is.
Repeating the pass without the BHC gives a second distribution
$\mathbf{p}^{-}_{i}$ at each position, and the contrast between the two forms
one of four feature categories: distributional, semantic, internal, and
medical knowledge.

We propose four types of sources that potentially affect the uncertainty: internal signals, medical knowledge, semantic, and distributional signals. To study how feature granularity affects performance, we define four configurations of increasing detail levels: 93, 120, 204, and up to 454 or 886 features. All models are evaluated on the 93, 120, and 204 configurations. We collect 454 as the maximum features for 7B and 8B models, while we collect 886 features for 70B models. This is because 70B models have 80 layers and therefore they need more layer-wise and attention features to sample. Complete details of all features are provided in Appendix~\ref{sec:features_details}.

\textbf{Internal Signals.} From the with-BHC pass, at each
sampled layer $l$, we extract the $\ell_2$ norm, mean, and the standard deviation of the
hidden state $\mathbf{h}^{(l)}_i$, together with the L2 distance and cosine
similarity between consecutive sampled layers.

Attention is characterized along three axes. Let
$\alpha^{(l,h)}_{i,j}$ be the weight from position $i$ to $j$ at layer $l$,
head $h$, and let $\mathcal{L}_{\mathrm{BHC}}$ index the BHC prefix.
\emph{Dispersion} is the entropy of the attention row:
\begin{equation}
H^{(l,h)}_i = -\sum_j \alpha^{(l,h)}_{i,j}\log \alpha^{(l,h)}_{i,j},
\end{equation}
distinguishing tokens that draw on a few specific positions from those whose
attention is diffuse. \emph{Grounding} is the total mass directed at the
clinical record:
\begin{equation}
A^{(l,h)}_i = \sum_{j\in\mathcal{L}_{\mathrm{BHC}}} \alpha^{(l,h)}_{i,j},
\end{equation}
which measures whether a token is supported by the record or only by the
surrounding summary. \emph{Stability} captures how much this grounding is
revised as representations are refined. A well-supported token should not
need to reallocate attention across depth. We measure it as the KL
divergence between the same head's attention rows at adjacent sampled
layers:
\begin{equation}
\mathrm{Drift}^{(h)}_{l\to l'}
= D_{\mathrm{KL}}\!\left(\tilde\alpha^{(l,h)}_{i}\,\big\|\,
\tilde\alpha^{(l',h)}_{i}\right),
\end{equation}
where $\tilde\alpha$ denotes the row after $\varepsilon$-smoothing and
renormalization.

To capture cumulative rather than per-layer grounding, we apply attention
rollout \citep{abnar2020quantifying}:
$\mathbf{R}_\ell=\prod_{k=0}^{\ell}(0.5\,\mathbf{A}_k+0.5\,\mathbf{I})$ with
row-wise renormalization at each step, and record the accumulated mass on
$\mathcal{L}_{\mathrm{BHC}}$, its entropy, and its maximum weight at a
shallow, a middle, and a deep checkpoint. This traces how far information
from the record actually reaches by each depth.

\textbf{Semantic Signals.} Given logits $\mathbf{z}$ and $\mathbf{p}=\mathrm{softmax}(\mathbf{z})$, we
compute Shannon entropy $H(\mathbf{p})=-\sum_v p_v\log p_v$, the top-two
margin $p_{(1)}-p_{(2)}$, the Gini coefficient of $\mathbf{p}$, and a
free-energy score $\mathcal{E}=-\log\sum_v \exp(z_v)$, which reflects the
overall evidence the model assigns at this position without softmax
normalization. Together these describe how concentrated or diffuse the prediction is. In traditional methods, uncertainty is considered to be high when the prediction is diffused.

\textbf{Medical Knowledge Signals.}
These encode domain context around each token. Named entity recognition is
run offline with scispaCy~\citep{scispacy} against UMLS/MedDRA
\citep{bodenreider2004umls}, yielding entity types and confidence scores.
Medical density measures the fraction of neighboring tokens recognized as
medical entities across window sizes. Corpus-level frequency and IDF are
computed over the full corpus. We also compute a conditional PMI, which
standardizes $\mathrm{PMI}_i$ within the entity type $e$ assigned to token
$i$:
\begin{equation}
    \mathrm{cPMI}_i = \frac{\mathrm{PMI}_i - \mu_e}{\sigma_e},
    \label{eq:cpmi}
\end{equation}
where $\mu_e$ and $\sigma_e$ are the mean and standard deviation of
$\mathrm{PMI}$ over all tokens of type $e$. These corpus-level statistics are
computed without reference to the annotations, so they do not leak label
information across the train/test split.

\textbf{Distributional Signals.}
The central quantity is the change the clinical record induces at each
position. We measure it as the point-wise mutual information between the
target token and the BHC,
\begin{equation}
    \mathrm{PMI}_i = \log p_{t_i} - \log p^{-}_{t_i},
    \label{eq:pmi}
\end{equation}
which is positive when the record raises the model's belief in the token,
and negative when the token is produced despite, rather than because of,
the record. We also use a normalized variant
$\mathrm{nPMI}_i = \mathrm{PMI}_i / (-\log p_{t_i})$, taking
$-\log p_{t_i}$ as the normalizer.

We compare the two distributions as a whole through $\Delta p$, $\Delta H$,
$\Delta \mathcal{E}$, and the KL and Jensen--Shannon divergences between
$\mathbf{p}$ and $\mathbf{p}^{-}$. Bringing in $\mathbf{p}^{\mathrm{prior}}$
separates the entropy reduction due to the record from that of the
surrounding text:
\begin{equation}
\begin{split}
    \underbrace{H(\mathbf{p}^{\mathrm{prior}}) - H(\mathbf{p}^{-})}_{\text{context}}
    &+ \underbrace{H(\mathbf{p}^{-}) - H(\mathbf{p})}_{\Delta H_{\mathrm{BHC}}} \\
    &= H(\mathbf{p}^{\mathrm{prior}}) - H(\mathbf{p}).
\end{split}
\label{eq:decomp}
\end{equation}
A token with $\Delta H_{\mathrm{BHC}} \approx 0$ is resolved by context
alone: fluent, but not grounded. Finally, neighborhood features aggregate
token probabilities over sliding windows around $i$, since tokens inside the
same unsupported span tend to share uncertainty patterns.

\subsection{Classifier}
We hypothesize that uncertainty is discretely distributed in the internal representations and that each model's internal activation differs accordingly. Therefore, fixed parameters, thresholds, and weights cannot accurately capture uncertainty, nor can they fairly generalize across different models. To address this, we adopt a supervised machine learning approach to train tree classifiers XGBoost~\cite{chen2016xgboost} and CatBoost~\cite{prokhorenkova2018catboost}, and linear classifier Logistic Regression~\cite{pedregosa2011scikit} on extracted features to predict token-level uncertainty according to the change of contexts.

Data is split at the case level, treating each BHC-summary pair as one case, with 80\% for training and 20\% for testing. Our method predicts a continuous uncertainty score in [0,1] for each token, and a threshold is then applied to determine whether a target token is flagged as uncertain. We then perform 5-fold cross-validation on test sets.  

\begin{table*}[t]
\centering
\resizebox{\textwidth}{!}{%
\begin{tabular}{lcccccccc}
\toprule
& \multicolumn{4}{c}{\textbf{MIMIC-DI}} & \multicolumn{4}{c}{\textbf{Generated-DI}} \\
\cmidrule(lr){2-5} \cmidrule(lr){6-9}
& \multicolumn{2}{c}{BioMistral-7B} & \multicolumn{2}{c}{OpenBioLLM-8B} & \multicolumn{2}{c}{BioMistral-7B} & \multicolumn{2}{c}{OpenBioLLM-8B} \\
\cmidrule(lr){2-3} \cmidrule(lr){4-5} \cmidrule(lr){6-7} \cmidrule(lr){8-9}
\textbf{Method} & AUCROC & AUPRC & AUCROC & AUPRC & AUCROC & AUPRC & AUCROC & AUPRC \\
\midrule
Token Entropy           & 0.4337 & 0.0658 & 0.4832 & 0.0767 & 0.4421 & 0.0366 & 0.5363 & 0.0440 \\
Semantic Energy         & 0.4714 & 0.0810 & 0.5375 & 0.0914 & 0.4697 & 0.0436 & 0.5534 & 0.0444 \\
Semantic Entropy        & 0.6448 & 0.0868 & 0.5804 & 0.0854 & 0.5361 & 0.1050 & 0.5475 & 0.0735 \\
Sliding Window Entropy  & 0.5127 & 0.0899 & 0.5877 & 0.1267 & 0.5884 & 0.0678 & 0.6940 & 0.1041 \\
Decomposing Uncertainty & 0.5938 & 0.0825 & 0.5657 & 0.0775 & 0.5925 & 0.1322 & 0.5959 & 0.1410 \\
SE Probes               & 0.7204 & 0.1313 & 0.6015 & 0.0587 & 0.4061 & 0.1075 & 0.5909 & 0.1553 \\
Uncertainty Aware Heads & 0.5508 & 0.0810 & 0.5488 & 0.0852 & 0.6150 & 0.0575 & 0.6403 & 0.0682 \\
Attention Chain         & 0.5195 & 0.0874 & 0.4913 & 0.1012 & 0.4549 & 0.0388 & 0.5098 & 0.0430 \\
\midrule
\textbf{Reverse Probing (XGBoost)}  & \textbf{0.8911} & \textbf{0.5478} & \underline{0.8833} & \underline{0.4976} & \textbf{0.8819} & \underline{0.2316} & \textbf{0.9364} & \underline{0.2892} \\
\textbf{Reverse Probing (CatBoost)} & \underline{0.8691} & \underline{0.5178} & \textbf{0.8848} & \textbf{0.5029} & \underline{0.8435} & \textbf{0.2805} & \underline{0.9144} & \textbf{0.2895} \\
\textbf{Reverse Probing (LRegression)} & 0.8508 & 0.3636 & 0.8545 & 0.4399 & 0.8122 & 0.1592 & 0.8807 & 0.2341 \\
\bottomrule
\end{tabular}%
}
\caption{Main results on MIMIC-DI and Generated-DI. Reverse Probing uses the max feature configuration (454 features). Baselines are adapted from general-domain methods as described in Section~\ref{sec:experiments}. Best results per column are \textbf{bold}; second best are \underline{underlined}.}
\label{tab:main_result}
\end{table*}

\subsection{Experiments}
\label{sec:experiments}
We evaluate on six LLMs: Mistral-7B-v0.1 \citep{jiang2023mistral}, Llama-3.1-8B-Instruct and Llama-3.1-70B-Instruct \citep{dubey2024llama3}, BioMistral-7B \citep{labrak2024biomistral}, Llama3-OpenBioLLM-8B and Llama3-OpenBioLLM-70B \citep{pal2024openbiollm}. The first three are general-domain models; the latter three are pretrained or fine-tuned on biomedical data.

No existing UQ method targets clinical summaries directly, so we adapted eight general-domain baselines to our setting: Token Entropy \citep{ma2025semantic}, Semantic Energy \citep{ma2025semantic}, Uncertainty Aware Heads \citep{vazhentsev2025uncertainty}, Attention Chain \citep{li2025uqac}, SE (Semantic Entropy) Probes \citep{kossen2024sep}, Sliding Window Entropy \citep{sriramanan2024llmcheck}, Decomposing Uncertainty \citep{hou2024decomposing}, and Semantic Entropy \citep{farquhar2024detecting,kuhn2023semantic}. Since our datasets consist of pre-existing clinical summaries rather than model-generated text, each baseline requires adaptation. Each adaptation preserves the core mechanism of the original method while replacing generation-dependent components with inference-only equivalents, ensuring a fair comparison as baselines.

For Semantic Entropy, we replace full-response sampling with sentence-level sampling: for each sentence in a summary, we prompt the LLM to explain it ten times, then compute semantic entropy over the resulting outputs. Sentences containing uncertain or hallucinated content tend to produce semantically divergent paraphrases, preserving the spirit of the original method.

For Sliding Window Entropy, we use the best hyperparameter configuration from grid search: window size 9, top-$k$ 20.

For SE Probes method, we use the sentence-level semantic entropy scores obtained from the Semantic Entropy adaptation above as supervision signal to train a linear probe over hidden states. We extract hidden states from every 4 layers in a single forward pass, covering shallow, middle, and deep representations. At test time, the probe predicts uncertainty from a single forward pass without any sampling, replacing the need for multiple generations. We use the same 80/20 train-test split as our other experiments. 

For baselines originally designed around masked token prediction, we replace masking with a cloze-style prompt: each token is withheld and the LLM predicts it from context, after which we compute Shannon entropy over the predicted distribution.

\section{Results and Analyses}

\subsection{Main Results}

We evaluate all methods at the token level. We report the AUPRC and AUCROC for
unsupported tokens across all documents. Micro F1 is reported in
Appendix~\ref{sec:complete_results}. Rather than judging whether an entire summary is reliable, we assess each token individually against expert annotations. While these metrics partially overlap with hallucination detection, our objective is fundamentally distinct and more intricate: we measure how models' internal representations activate when encountering unsupported facts.

\begin{table}[t]
\centering
\small
\setlength{\tabcolsep}{4pt}
\begin{tabular}{llcc}
\toprule
\textbf{Model} & \textbf{Best baseline} & \textbf{Baseline} & \textbf{RP} \\
\midrule
\multicolumn{4}{l}{\textit{MIMIC-DI}} \\
Mistral-7B      & Attention Chain & 0.0918 & \textbf{0.6022} \\
BioMistral-7B   & SE Probes       & 0.1313 & 0.5478 \\
Llama3.1-8B     & SW Entropy             & 0.1171 & 0.5140 \\
OpenBioLLM-8B   & SW Entropy             & 0.1267 & 0.5029 \\
Llama3.1-70B    & SW Entropy             & 0.1090 & 0.3934 \\
OpenBioLLM-70B  & SW Entropy             & 0.1244 & 0.3597 \\
\midrule
\multicolumn{4}{l}{\textit{Generated-DI}} \\
Mistral-7B      & SE Probes       & 0.2067 & \textbf{0.3070} \\
BioMistral-7B   & Decomp.\ Unc.   & 0.1322 & 0.2805 \\
Llama3.1-8B     & SE Probes       & 0.1386 & 0.2761 \\
OpenBioLLM-8B   & SE Probes       & 0.1553 & 0.2895 \\
Llama3.1-70B    & SE Probes       & 0.1830 & 0.1939 \\
OpenBioLLM-70B  & SE Probes       & 0.1709 & 0.2073 \\
\bottomrule
\end{tabular}
\caption{\textbf{Best baseline} is the best result among all
eight baselines for that model and dataset (SW Entropy refers to Sliding Window Entropy),
with its AUPRC in \textbf{Baseline}; \textbf{RP} is Reverse Probing AUPRC (best classifier, max feature configuration). Best result per dataset in \textbf{bold}.}
\label{tab:scale}
\end{table}

Although our AUCROC values are high, this are partially affected by the severe class imbalance in both datasets: 7.3\% unsupported tokens in MIMIC-DI, and only 2.05\% unsupported tokens in Generated-DI. AUCROC inflates because the true negatives dominate the score. AUPRC is more informative here, so we utilize it as the main matric. Our results show consistent improvement on this metric across all models and configurations. Among all models and configurations, our two tree-structured classifiers XGBoost and CatBoost perform comparably.  Logistic regression is consistently weaker. However, it still beats every baseline on all six models on MIMIC-DI, which shows that the features themselves has already carried the signal. Comparing to linear models, tree classifiers perform better, which suggest the uncertainty has complex feature interactions.

Table~\ref{tab:main_result} reports results of BioMistral-7B and OpenBioLlama-8B on AUCROC and AUPRC for all methods on both datasets. The two models are pretrained by medical data. Table~\ref{tab:scale} reports Reverse Probing AUPRC across model scale and pretraining condition. Reverse Probing outperforms all baselines by a large margin on both metrics. On MIMIC-DI, where 7.3\% of tokens are unsupported, it reaches 0.6022 AUPRC with Mistral-7B against 0.0918 for the strongest baseline on that model, and the margin ranges from $2.9\times$ to $6.6\times$ across the six models. XGBoost achieves an AUPRC of 0.5478 with BioMistral-7B, compared to 0.1313 for the strongest baseline (SE Probes), a roughly $4\times$ improvement. The gap is smaller but consistent on Generated-DI, where positive cases are only 2.1\%, and all methods score lower on all metrics.

Most of the baselines perform poorly. On MIMIC-DI, the strongest of the eight reaches 0.13 AUPRC against a random baseline of 0.07, and several fall below 0.5 AUCROC, meaning their scores rank unsupported tokens no better than chance and sometimes in the wrong direction. The gap between baselines and Reverse Probing reflects both the difficulty of clinical text and the advantage of token-level features extracted directly from internal model signals.

From efficiency point of view, Reverse Probing requires only two forward passes per summary: with and without BHC. This method executes 100 summaries in under 20 minutes on a single B200 GPU, compared to an average of 7 hours for multi-sampling-based baselines such as Semantic Entropy and SE Probes. This efficiency comes from operating directly on pre-existing text with only predicting one masked token each time, making it a practical candidate for real-time clinical deployment. Table~\ref{tab:efficiency} in Appendix~\ref{app:efficiency} reports the full comparison.

\begin{comment}

\begin{figure}[t]
    \centering
    \includegraphics[width=\linewidth]{images/robustness_dense_bar.png}
    \caption{AUPRC across base LLMs for four selected baselines and Reverse Probing. Each bar represents one base LLM using the max feature configuration (454 for 7-8B models and 886 for 70B models).}
    \label{fig:more_comparison}
\end{figure}

Figure~\ref{fig:more_comparison} extends this comparison to all six base LLMs. Reverse Probing maintains a clear and consistent advantage across models of varying size and biomedical pretraining, while baseline performance varies substantially across LLMs. This suggests that Reverse Probing is robust to different base models.
\end{comment}

\begin{figure}[t]
    \centering
    \includegraphics[width=0.9\linewidth]{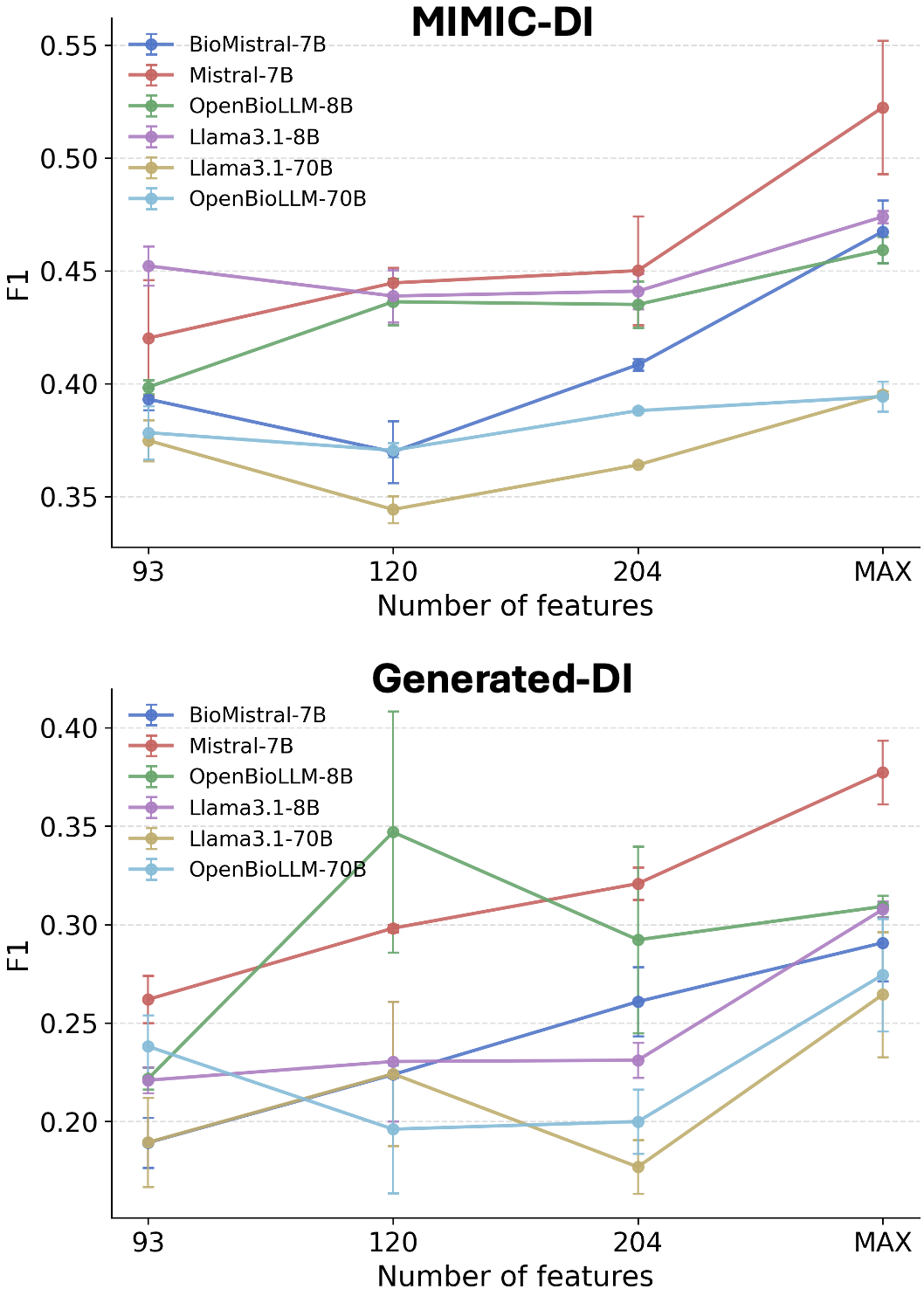}
    \caption{F1 vs.\ number of features on MIMIC-DI and Generated-DI. Each point is the mean over XGBoost and CatBoost. MAX
denotes 454 features for 7/8B models and 886 for 70B models.}
    \label{fig:feature_number}
\end{figure}

\begin{figure*}[t]
    \centering
    \includegraphics[width=1\linewidth]{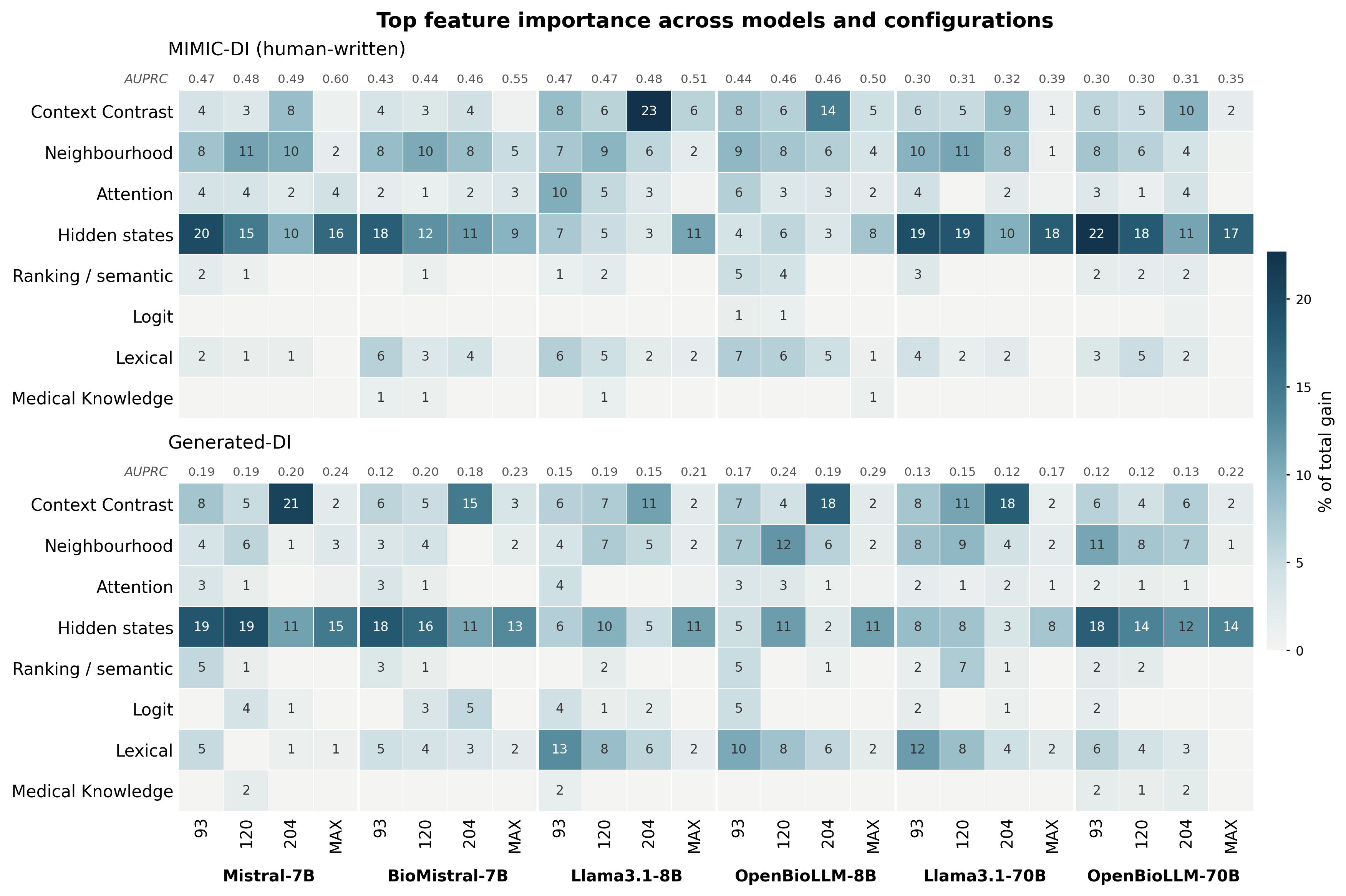}
    \caption{Feature importance across models and feature configurations, using
XGBoost. Rows group features into the eight families of
Section~\ref{sec:features}, which map onto four signal categories:
distributional (context contrast, neighborhood), semantic (logit, ranking \&
semantic), internal (hidden states, attention), and medical knowledge
(medical NER, lexical). Each column is one (model, configuration) pair, and
each cell gives the share of total gain carried by that family's features
among the top 20. Darker cells carry a larger share. AUPRC for each configuration is
shown above the column. MAX denotes 454 features for 7/8B models and 886 for
70B models.}
    \label{fig:heatmap}
\end{figure*}

\begin{figure}[ht]
    \centering
    \includegraphics[width=1\linewidth]{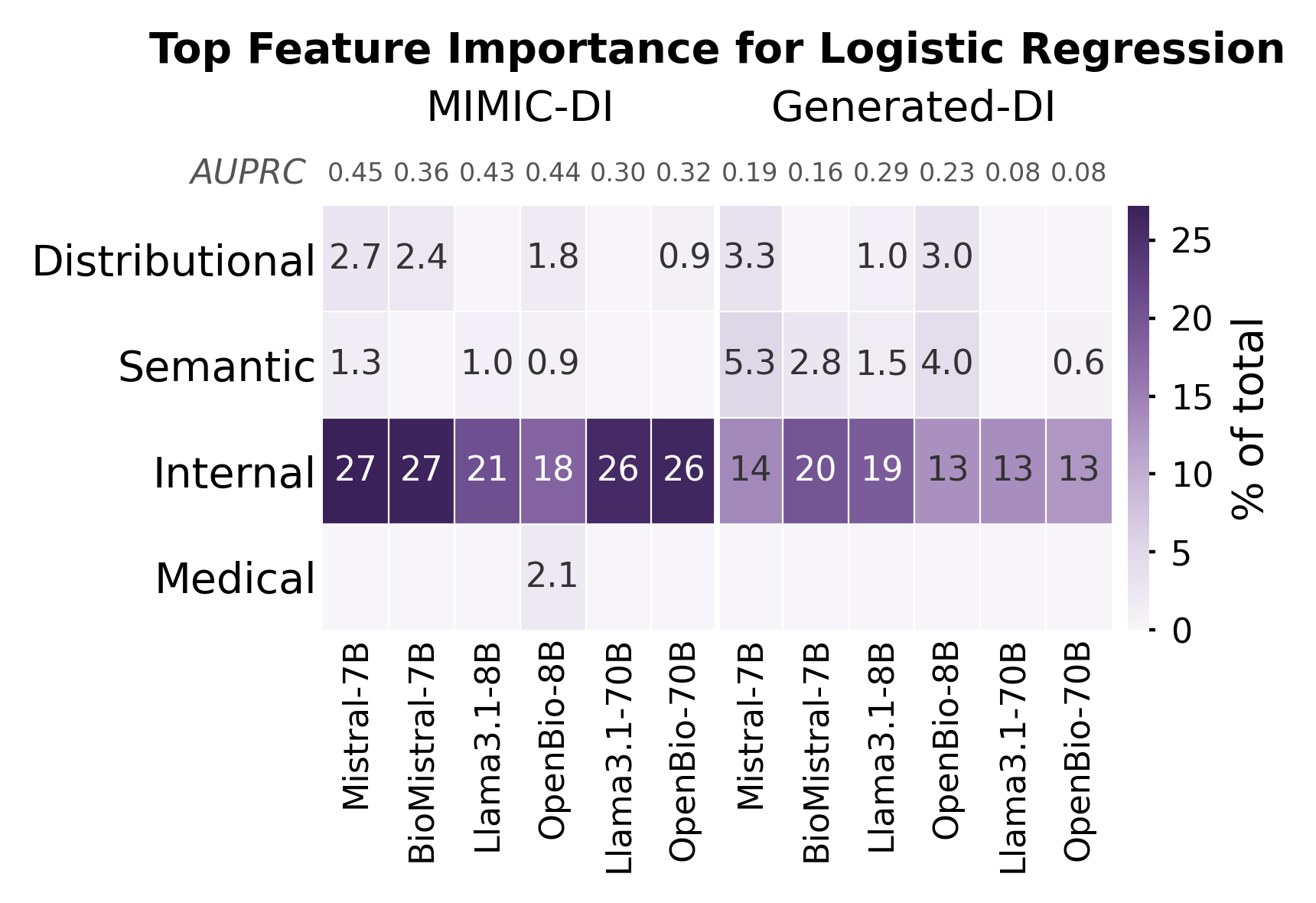}
    \caption{Top 20 recurring features by importance across models on MAX features, using Logistic Regression classifier, with contribution of the top 20 features and AUPRC above each column. AUPRC is shown above each column.}
    \label{fig:liear_classifier}
\end{figure}

An unexpected finding is that general-purpose models such as Mistral-7B and Llama3.1-8B sometimes outperform biomedical models such as OpenBioLLM-8B. The best result of MIMIC-DI and Generated-DI also comes from Mistral-7B, which is a smaller general domain model. An explanation is that biomedical pretraining makes models more confident in medical content, suppressing the uncertainty signals that our features rely on. An overconfident model produces flatter entropy differences and smaller delta energy, making uncertain tokens harder to distinguish from supported ones.

The results of 70B models are also lower than those of smaller models. We hypothesize that larger models tend to produce overconfident predictions across all contexts, compressing the dynamic range of uncertainty features and making it harder for the classifier to separate unsupported from supported tokens. Additionally, larger models have more layers, and uncertainty signals may be more diffusely distributed across their representations. However, with a larger number of model layers, the sampling is effectively sparser for large models when the number of sampled features remains the same. Our current feature set samples a fixed number of layers, which may not capture enough information in 70B models compared to smaller models. This is consistent with findings in the literature showing that model scale does not reliably improve uncertainty estimation \citep{guo2017calibration}.

\subsection{Uncertainty Source}

Figure~\ref{fig:feature_number} shows how F1 changes with collected feature number increased on
MIMIC-DI and Generated-DI. Performance generally increases with more
features.
From 93 to 120 features, we added more NER-derived features with scispaCy, and add a wider neighbourhood window ($w=7$). Despite the larger feature
count, F1 drops slightly for several models at this stage, suggesting that
the NER-derived entity-type and confidence signals may be noisier. Performance recovers
at 204 features, while the additional
context-contrast, PMI, entropy-decomposition, and medical-prior-decomposition
features are added. Given that more features enhance the final result, this aligns with our hypothesis that uncertainty exists in a discrete form within the internal representations of models.

Figure~\ref{fig:heatmap} shows the most consistently important features across models and configurations from XGBoost. Of the four categories we defined, only two carry
substantial weight: distributional signals and internal signals. Under XGBoost, \texttt{delta\_energy} enters the top 20 in 47 of the 48 configurations and averages 2.65\% of total gain, the highest of any single feature. The second is \texttt{neighbor\_avg\_w5} at 1.69\%. CatBoost gives the similar result: \texttt{delta\_energy} appears in all 48 configurations
with a mean share of 5.12\%, close to twice the second-ranked feature \texttt{neighbor\_avg\_w5} (2.67\%). The BHC context difference therefore holds up as the most consistent predictor across two classifiers and both datasets.

Neighborhood features rank second and stay stable across configurations. This matches prior work showing that sliding window entropy improves uncertainty estimation by incorporating local context \citep{sriramanan2024llmcheck}, and it fits the structure of clinical text, where medical concepts span several tokens. The uncertainty of a single token rarely means much on its own, and neighborhood signals let the classifier judge whether a token sits inside a broader uncertain region.

The two datasets differ in emphasis rather than in kind. Neighborhood features weigh more heavily in MIMIC-DI. This suggests that uncertainty in human-written notes clusters within local spans. Context contrast features like delta energy weigh more heavily in Generated-DI. This indicates generated content tracks how the BHC shifts the output distribution more closely.

Only the tree classifiers utilized the distributional signals. In Figure~\ref{fig:liear_classifier}, Logistic regression places almost none of its weight on distributional signals, and it concentrates on internal signals. Within that group the weight concentrates on layer-wise norms and layer-to-layer change. This shows capturing the distributional signals requires a model that can represent interactions between features rather than weight them independently.

This finding also helps explain why traditional UQ methods underperformed. Methods such as Token Entropy and Semantic Energy measure the output distribution at the final layer, which is categorized as Semantic Signals in our feature extraction process. However, in our final result, Semantic Signals contribute the least among all classifiers. Traditional methods missed the shift induced by removing the BHC, and the local structure that ties neighboring tokens together. Our results suggest that these deeper internal signals carry most of the useful uncertainty information in clinical text. 

% Required packages: \usepackage{booktabs,multirow,makecell}

% Required packages: \usepackage{booktabs,multirow,makecell}

% Required packages: \usepackage{booktabs,multirow,makecell}

% Required packages:
% \usepackage{booktabs}
% \usepackage{graphicx}

% Required packages in preamble:
% \usepackage{booktabs}
% \usepackage{graphicx}  % for \resizebox

% Required packages in preamble:
% \usepackage{booktabs}
% \usepackage{graphicx}  % for \resizebox

\begin{comment}
\begin{figure}[h]
    \centering
    \includegraphics[width=\linewidth]{images/alignment.png}
    \caption{Validation framework for UQ metric evaluation. The proposed framework compares UQ metrics against external benchmark results using Kendall's $\tau$, evaluated through intragroup, inter-method, and external correlation analyses.}
    \label{fig:alignment}
\end{figure}
\end{comment}

\section{Conclusion}

We introduce Reverse Probing, the first token-level uncertainty quantification framework for clinical text. Unlike traditional UQ methods that rely on generating new text, we show that pre-existing clinical summaries can serve as probes into the model's internal state. By feeding existing text into a frozen LLM and taking snapshots of its internal representations, we train a classifier to detect whether hidden states and output distributions react to unsupported content. This eliminates text generation entirely, reducing inference time from 7-8 hours to under 20 minutes, and making uncertainty estimation practical where evaluating full generated summaries is infeasible. Reverse Probing consistently outperforms all adapted baselines, achieving approximately 2.9 to 6.6 times AUPRC improvement over all strongest baselines.

Feature analysis reveals that delta energy and neighborhood signals are the most stable uncertainty predictors. Delta energy captures how much BHC context shifts the entire output distribution. Neighborhood features reflect a fundamental property of clinical text: medical concepts span multiple tokens, and uncertainty rarely appears in isolation. These signals suggest that LLMs internally encode uncertainty in response to unsupported content. However, the uncertainty is distributed diffusely and interactively. By tracing predictions back to specific internal representations, Reverse Probing turns uncertainty estimation into an interpretable process, taking a step toward explainable AI in clinical settings.

% \section*{Acknowledgments}

\section*{Limitations}
\textbf{Low-resource data.} Our evaluation is limited to two datasets, both derived from MIMIC-IV. To our knowledge, these are the only publicly available clinical summarization corpora with span-level annotations of unsupported facts. This kind of annotation requires trained clinicians to verify each claim against the source record, which is slow, costly, and difficult to scale.

\textbf{Large model adaptability.} Our framework performs better on 7-8B models than to 70B models. Larger models have more layers and more diffuse internal representations, and we do not design a targeted feature extraction strategy for this scale. Adaptive layer sampling for large models is a natural direction for future work.

\textbf{Single task generalization.} We evaluate exclusively on discharge summaries, and it remains to be seen whether Reverse Probing generalizes to other clinical text types such as radiology reports or progress notes.

\textbf{Uncertainty categorization.} While our feature analysis traces uncertainty back to its source signals, we do not categorize uncertainty by type. The original datasets provide eleven fine-grained labels for unsupported facts, and it remains an open question whether different uncertainty sources systematically correspond to different error types. Linking feature-level signals to semantic categories of unsupported content would be a meaningful step toward deeper interpretability.

\section*{Ethical Considerations}
This work uses Hallucinations-MIMIC-DI and Hallucinations-Generated-DI introduced by \citet{Hegselmann}. Both datasets derive from MIMIC-IV-Note \citep{PhysioNet-mimic-iv-note-2.2}, a de-identified clinical corpus. Access to MIMIC-IV-Note requires credentialed registration, completion of CITI Data or Specimens Only Research training, and a signed data use agreement through PhysioNet. All authors met these requirements before accessing any data, in compliance with the PhysioNet data use policy. No new patient data were collected. Our framework is a research tool for evaluating model uncertainty against existing expert annotations.

\bibliography{custom}

@article{farquhar2024detecting,
  title={Detecting hallucinations in large language models using semantic entropy},
  author={Farquhar, Sebastian and Kossen, Jannik and Kuhn, Lorenz and Gal, Yarin},
  journal={Nature},
  volume={630},
  number={8017},
  pages={625--630},
  year={2024},
  publisher={Nature Publishing Group UK London}
}

@inproceedings{
manakul2023selfcheckgpt,
title={SelfCheck{GPT}: Zero-Resource Black-Box Hallucination Detection for Generative Large Language Models},
author={Potsawee Manakul and Adian Liusie and Mark Gales},
booktitle={The 2023 Conference on Empirical Methods in Natural Language Processing},
year={2023},
url={https://openreview.net/forum?id=RwzFNbJ3Ez}
}

@inproceedings{
kuhn2023semantic,
title={Semantic Uncertainty: Linguistic Invariances for Uncertainty Estimation in Natural Language Generation},
author={Lorenz Kuhn and Yarin Gal and Sebastian Farquhar},
booktitle={The Eleventh International Conference on Learning Representations },
year={2023},
url={https://openreview.net/forum?id=VD-AYtP0dve}
}

@inproceedings{hou2024decomposing,
author = {Hou, Bairu and Liu, Yujian and Qian, Kaizhi and Andreas, Jacob and Chang, Shiyu and Zhang, Yang},
title = {Decomposing uncertainty for large language models through input clarification ensembling},
year = {2024},
publisher = {JMLR.org},
booktitle = {Proceedings of the 41st International Conference on Machine Learning},
articleno = {765},
numpages = {20},
location = {Vienna, Austria},
series = {ICML'24}
}

@misc{liu2024uncertaintyestimationquantificationllms,
      title={Uncertainty Estimation and Quantification for LLMs: A Simple Supervised Approach}, 
      author={Linyu Liu and Yu Pan and Xiaocheng Li and Guanting Chen},
      year={2024},
      eprint={2404.15993},
      archivePrefix={arXiv},
      primaryClass={cs.LG},
      url={https://arxiv.org/abs/2404.15993}, 
}

@inproceedings{guo2017calibration,
  title={On calibration of modern neural networks},
  author={Guo, Chuan and Pleiss, Geoff and Sun, Yu and Weinberger, Kilian Q},
  booktitle={International conference on machine learning},
  pages={1321--1330},
  year={2017},
  organization={PMLR}
}

@article{lundberg2017unified,
  title={A unified approach to interpreting model predictions},
  author={Lundberg, Scott M and Lee, Su-In},
  journal={Advances in neural information processing systems},
  volume={30},
  year={2017}
}

@article{qureshi2025explainability,
  title={Explainability in action: A metric-driven assessment of local explanations for healthcare tabular models},
  author={Qureshi, M Atif and Noor, Abdul Aziz and Manzoor, Awais and Mazhar Qureshi, Muhammad Deedahwar and Younus, Arjumand and Rashwan, Wael},
  journal={medRxiv},
  pages={2025--05},
  year={2025},
  publisher={Cold Spring Harbor Laboratory Press}
}

@article{hur2025comparison,
  title={Comparison of SHAP and clinician friendly explanations reveals effects on clinical decision behaviour},
  author={Hur, Sujeong and Lee, Yura and Park, Joongheum and Jeon, Yeong Jeong and Cho, Jong Ho and Cho, Duck and Lim, Dobin and Hwang, Wonil and Cha, Won Chul and Yoo, Junsang},
  journal={NPJ Digital Medicine},
  volume={8},
  number={1},
  pages={578},
  year={2025},
  publisher={Nature Publishing Group UK London}
}

@article{salvi2025explainability,
  title={Explainability and uncertainty: Two sides of the same coin for enhancing the interpretability of deep learning models in healthcare},
  author={Salvi, Massimo and Seoni, Silvia and Campagner, Andrea and Gertych, Arkadiusz and Acharya, U Rajendra and Molinari, Filippo and Cabitza, Federico},
  journal={International Journal of Medical Informatics},
  volume={197},
  pages={105846},
  year={2025},
  publisher={Elsevier}
}

@article{dubey2025ubiqtree,
  title={UbiQTree: Uncertainty quantification in XAI with tree ensembles},
  author={Dubey, Akshat and An{\v{z}}el, Aleksandar and {\.I}lgen, Bahar and Hattab, Georges},
  journal={Patterns},
  year={2025},
  publisher={Elsevier}
}

@article{lu2024uncertainty,
  title={Uncertainty quantification and interpretability for clinical trial approval prediction},
  author={Lu, Yingzhou and Chen, Tianyi and Hao, Nan and Van Rechem, Capucine and Chen, Jintai and Fu, Tianfan},
  journal={Health Data Science},
  volume={4},
  pages={0126},
  year={2024},
  publisher={AAAS}
}

@inproceedings{adams2023meta,
  title={A meta-evaluation of faithfulness metrics for long-form hospital-course summarization},
  author={Adams, Griffin and Zuckerg, Jason and Elhadad, No{\'e}mie},
  booktitle={Machine Learning for Healthcare Conference},
  pages={2--30},
  year={2023},
  organization={PMLR}
}

@inproceedings{xie2024doclens,
  title={Doclens: Multi-aspect fine-grained medical text evaluation},
  author={Xie, Yiqing and Zhang, Sheng and Cheng, Hao and Liu, Pengfei and Gero, Zelalem and Wong, Cliff and Naumann, Tristan and Poon, Hoifung and Rose, Carolyn},
  booktitle={Proceedings of the 62nd Annual Meeting of the Association for Computational Linguistics (Volume 1: Long Papers)},
  pages={649--679},
  year={2024}
}

@article{croxford2025evaluating,
  title={Evaluating clinical AI summaries with large language models as judges},
  author={Croxford, Emma and Gao, Yanjun and First, Elliot and Pellegrino, Nicholas and Schnier, Miranda and Caskey, John and Oguss, Madeline and Wills, Graham and Chen, Guanhua and Dligach, Dmitriy and others},
  journal={npj Digital Medicine},
  volume={8},
  number={1},
  pages={640},
  year={2025},
  publisher={Nature Publishing Group UK London}
}

@inproceedings{wang2025semantic,
  title={Semantic consistency-based uncertainty quantification for factuality in radiology report generation},
  author={Wang, Chenyu and Zhou, Weichao and Ghosh, Shantanu and Batmanghelich, Kayhan and Li, Wenchao},
  booktitle={Findings of the Association for Computational Linguistics: NAACL 2025},
  pages={1739--1754},
  year={2025}
}

@article{PhysioNet-mimic-iv-note-2.2,
  author = {Johnson, Alistair and Pollard, Tom and Horng, Steven and Celi, Leo Anthony and Mark, Roger},
  title = {{MIMIC-IV-Note: Deidentified free-text clinical notes}},
  journal = {{PhysioNet}},
  year = {2023},
  month = jan,
  note = {Version 2.2},
  doi = {10.13026/1n74-ne17},
  url = {https://doi.org/10.13026/1n74-ne17}
}

@article{Hegselmann,
  author = {Hegselmann, Stefan and Shen, Shannon and Gierse, Florian and Agrawal, Monica and Sontag, David and Jiang, Xiaoyi},
  title = {{Medical Expert Annotations of Unsupported Facts in Doctor-Written and LLM-Generated Patient Summaries}},
  journal = {{PhysioNet}},
  year = {2025},
  month = apr,
  note = {Version 1.0.1},
  doi = {10.13026/gedc-j464},
  url = {https://doi.org/10.13026/gedc-j464}
}

@inproceedings{hegselmann2024data,
  title={A Data-Centric Approach To Generate Faithful and High Quality Patient Summaries with Large Language Models},
  author={Hegselmann, Stefan and Shen, Zejiang and Gierse, Florian and Agrawal, Monica and Sontag, David and Jiang, Xiaoyi},
  booktitle={Conference on Health, Inference, and Learning},
  pages={339--379},
  year={2024},
  organization={PMLR}
}

@inproceedings{scispacy,
    title = "{S}cispa{C}y: Fast and Robust Models for Biomedical Natural Language Processing",
    author = "Neumann, Mark  and
      King, Daniel  and
      Beltagy, Iz  and
      Ammar, Waleed",
    editor = "Demner-Fushman, Dina  and
      Cohen, Kevin Bretonnel  and
      Ananiadou, Sophia  and
      Tsujii, Junichi",
    booktitle = "Proceedings of the 18th BioNLP Workshop and Shared Task",
    month = aug,
    year = "2019",
    address = "Florence, Italy",
    publisher = "Association for Computational Linguistics",
    url = "https://aclanthology.org/W19-5034/",
    doi = "10.18653/v1/W19-5034",
    pages = "319--327"
}

@article{bodenreider2004umls,
  author  = {Bodenreider, Olivier},
  title   = {The Unified Medical Language System ({UMLS}): integrating biomedical terminology},
  journal = {Nucleic Acids Research},
  year    = {2004},
  volume  = {32},
  number  = {Database issue},
  pages   = {D267--D270},
  doi     = {10.1093/nar/gkh061},
  pmid    = {14681409},
  pmcid   = {PMC308795},
}

@misc{yona2026,
      title={Hallucinations Undermine Trust; Metacognition is a Way Forward}, 
      author={Gal Yona and Mor Geva and Yossi Matias},
      year={2026},
      eprint={2605.01428},
      archivePrefix={arXiv},
      primaryClass={cs.CL},
      url={https://arxiv.org/abs/2605.01428}, 
}

@inproceedings{chen2016xgboost,
  title     = {{XGBoost}: A Scalable Tree Boosting System},
  author    = {Chen, Tianqi and Guestrin, Carlos},
  booktitle = {Proceedings of the 22nd ACM SIGKDD International Conference
               on Knowledge Discovery and Data Mining},
  pages     = {785--794},
  year      = {2016},
  doi       = {10.1145/2939672.2939785}
}

@inproceedings{prokhorenkova2018catboost,
  title     = {{CatBoost}: Unbiased Boosting with Categorical Features},
  author    = {Prokhorenkova, Liudmila and Gusev, Gleb and Vorobev, Aleksandr
               and Dorogush, Anna Veronika and Gulin, Andrey},
  booktitle = {Advances in Neural Information Processing Systems},
  volume    = {31},
  pages     = {6639--6649},
  year      = {2018}
}

@article{ma2025semantic,
  title   = {Semantic Energy: Detecting {LLM} Hallucination Beyond Entropy},
  author  = {Ma, Hao and Pan, Jiacheng and Liu, Jiafei and Chen, Yuchen and
             Zhou, Joey Tianyi and Wang, Gang and others},
  journal = {arXiv preprint arXiv:2508.14496},
  year    = {2025}
}

@article{vazhentsev2025uncertainty,
  title   = {Uncertainty-Aware Attention Heads: Efficient Unsupervised
             Uncertainty Quantification for {LLMs}},
  author  = {Vazhentsev, Artem and Rvanova, Lyudmila and Kuzmin, Gleb and
             Fadeeva, Ekaterina and Lazichny, Ivan and Panchenko, Alexander and
             Panov, Maxim and Baldwin, Timothy and Sachan, Mrinmaya and
             Nakov, Preslav and Shelmanov, Artem},
  journal = {arXiv preprint arXiv:2505.20045},
  year    = {2025}
}

@inproceedings{li2025uqac,
  title     = {Language Model Uncertainty Quantification with Attention Chain},
  author    = {Li, Yinghao and Qiang, Rushi and Moukheiber, Lama and Zhang, Chao},
  booktitle = {Proceedings of the Conference on Language Modeling (COLM)},
  year      = {2025},
  note      = {arXiv:2503.19168}
}

@inproceedings{kossen2024sep,
  title     = {Semantic Entropy Probes: Robust and Cheap Hallucination Detection in {LLMs}},
  author    = {Kossen, Jannik and Han, Jiatong and Razzak, Muhammed and
               Schut, Lisa and Malik, Shreshth and Gal, Yarin},
  booktitle = {Proceedings of the International Conference on Machine Learning (ICML)},
  year      = {2024},
  note      = {arXiv:2406.15927}
}

@inproceedings{sriramanan2024llmcheck,
  title     = {{LLM-Check}: Investigating Detection of Hallucinations in
               Large Language Models},
  author    = {Sriramanan, Ganesh and Bharti, Sumit and Sadasivan, Vinu Sankar and
               Saha, Srivatsa and Kattakinda, Priyatham and Feizi, Soheil},
  booktitle = {Advances in Neural Information Processing Systems},
  volume    = {37},
  pages     = {34188--34216},
  year      = {2024}
}

@article{jiang2023mistral,
  title   = {Mistral {7B}},
  author  = {Jiang, Albert Q. and Sablayrolles, Alexandre and Mensch, Arthur
             and Bamford, Chris and Chaplot, Devendra Singh and
             de las Casas, Diego and Bressand, Florian and Lengyel, Gianna
             and Lample, Guillaume and Saulnier, Lucile and Lavaud, L{\'e}lio Renard
             and Lachaux, Marie-Anne and Stock, Pierre and Scao, Teven Le
             and Lavril, Thibaut and Wang, Thomas and Lacroix, Timoth{\'e}e
             and Sayed, William El},
  journal = {arXiv preprint arXiv:2310.06825},
  year    = {2023}
}

@article{dubey2024llama3,
  title   = {The {Llama} 3 Herd of Models},
  author  = {Dubey, Abhimanyu and Jauhri, Abhinav and Pandey, Abhinav and others},
  journal = {arXiv preprint arXiv:2407.21783},
  year    = {2024}
}

@misc{labrak2024biomistral,
  title         = {{BioMistral}: A Collection of Open-Source Pretrained Large
                   Language Models for Medical Domains},
  author        = {Labrak, Yanis and Bazoge, Adrien and Morin, Emmanuel and
                   Gourraud, Pierre-Antoine and Rouvier, Mickael and Dufour, Richard},
  year          = {2024},
  eprint        = {2402.10373},
  archivePrefix = {arXiv},
  primaryClass  = {cs.CL}
}

@misc{pal2024openbiollm,
  title        = {{OpenBioLLMs}: Advancing Open-Source Large Language Models
                  for Healthcare and Life Sciences},
  author       = {Pal, Ankit and Sankarasubbu, Malaikannan},
  year         = {2024},
  publisher    = {Hugging Face},
  howpublished = {\url{https://huggingface.co/aaditya/OpenBioLLM-Llama3-70B}}
}

@inproceedings{devlin-etal-2019-bert,
    title = "{BERT}: Pre-training of Deep Bidirectional Transformers for Language Understanding",
    author = "Devlin, Jacob  and
      Chang, Ming-Wei  and
      Lee, Kenton  and
      Toutanova, Kristina",
    editor = "Burstein, Jill  and
      Doran, Christy  and
      Solorio, Thamar",
    booktitle = "Proceedings of the 2019 Conference of the North {A}merican Chapter of the Association for Computational Linguistics: Human Language Technologies, Volume 1 (Long and Short Papers)",
    month = jun,
    year = "2019",
    address = "Minneapolis, Minnesota",
    publisher = "Association for Computational Linguistics",
    url = "https://aclanthology.org/N19-1423/",
    doi = "10.18653/v1/N19-1423",
    pages = "4171--4186"
}

@inproceedings{huang2026detecting,
title={Detecting Misbehaviors of Large Vision-Language Models by Evidential Uncertainty Quantification},
author={Tao Huang and Rui Wang and Xiaofei Liu and Yi Qin and Li Duan and Liping Jing},
booktitle={The Fourteenth International Conference on Learning Representations},
year={2026},
url={https://openreview.net/forum?id=xJT4fXJr1Q}
}

@inproceedings{abnar2020quantifying,
  title={Quantifying attention flow in transformers},
  author={Abnar, Samira and Zuidema, Willem},
  booktitle={Proceedings of the 58th annual meeting of the association for computational linguistics},
  pages={4190--4197},
  year={2020}
}

@article{pedregosa2011scikit,
  title={Scikit-learn: Machine learning in Python},
  author={Pedregosa, Fabian and Varoquaux, Ga{\"e}l and Gramfort, Alexandre and Michel, Vincent and Thirion, Bertrand and Grisel, Olivier and Blondel, Mathieu and Prettenhofer, Peter and Weiss, Ron and Dubourg, Vincent and others},
  journal={the Journal of machine Learning research},
  volume={12},
  pages={2825--2830},
  year={2011},
  publisher={JMLR. org}
}

\appendix

\section{Implementation Details}
\label{app:efficiency}
All experiments are conducted on a single NVIDIA
B200 GPU. Table~\ref{tab:efficiency} breaks down the cost. Reverse Probing requires only two
forward passes per summary and processes 100
clinical summaries in less than 20 minutes.
Methods requiring multiple sampling passes are more expensive: Semantic Entropy,
Decomposing Uncertainty, and SE Probes each
require 7-8 hours for the same 100 samples, and
Uncertainty Aware Heads takes approximately 1.5
hours. Token Entropy, Semantic Energy, Sliding
Window Entropy, and Attention Chain complete
within a few minutes, comparable to our method,
but performance is much lower.

\begin{table}[h]
\centering
\small
\begin{tabular}{l c c}
\toprule
\textbf{Method} & \textbf{Fwd.\ passes} & \textbf{Time} \\
\midrule
\multicolumn{3}{l}{\textit{Sampling-based}} \\
Semantic Entropy        & $S \times N_s$ & 7--8\,h \\
Decomposing Uncertainty & $S \times N_s$ & 7--8\,h \\
SE Probes               & $S \times N_s$ & 7--8\,h \\
\midrule
\multicolumn{3}{l}{\textit{Per-token prompting}} \\
Uncertainty Aware Heads & $W$ & 1.5\,h \\
\midrule
\multicolumn{3}{l}{\textit{Inference-only}} \\
Token Entropy           & 1 & 5 min \\
Semantic Energy         & 1 & 5 min \\
Sliding Window Entropy  & 1 & 10 min \\
Attention Chain         & 1 & 10 min \\
\midrule
\textbf{Reverse Probing} & \textbf{2} & \textbf{15+2 min} \\
\bottomrule
\end{tabular}
\caption{Processing time of 100 clinical summaries on a single
NVIDIA B200. $N_s$ is the number of sampled generations per sentence, $S$ is
the number of sentences per summary and $W$ is the number of words per summary. The 15 minutes for Reverse Probing denotes the time cost for feature collection, and the 2 minutes denotes the time for training and testing.}
\label{tab:efficiency}
\end{table}

\section{Example Result}

\label{sec:result_example}
\begin{figure}[h]
    \centering
    \includegraphics[width=\linewidth]{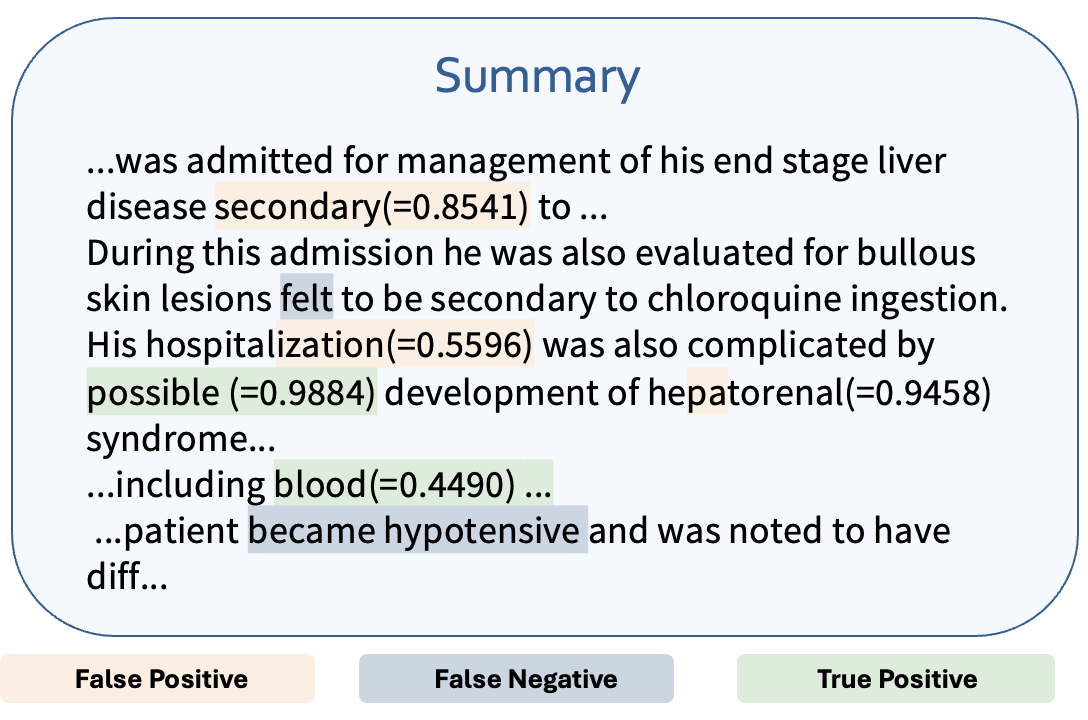}
    \caption{The ground truth labels and uncertainty quantification results for a sample clinical summary.}
    \label{fig:example_result}
\end{figure}

Figure~\ref{fig:example_result} shows a sample output of Reverse Probing on a
clinical summary. Tokens flagged with uncertainty scores are highlighted in three
colors: true positives (green) correctly identify tokens that are factually
unsupported, false positives (red) are tokens flagged but not annotated as
uncertain, and false negatives (blue) are annotated uncertain tokens that the
model missed. The uncertainty score shown next to each flagged token reflects
the classifier's estimation that the model is not sure about this token.

\section{CatBoost Feature Contributions}
\begin{figure*}[h]
    \centering
    \includegraphics[width=1\linewidth]{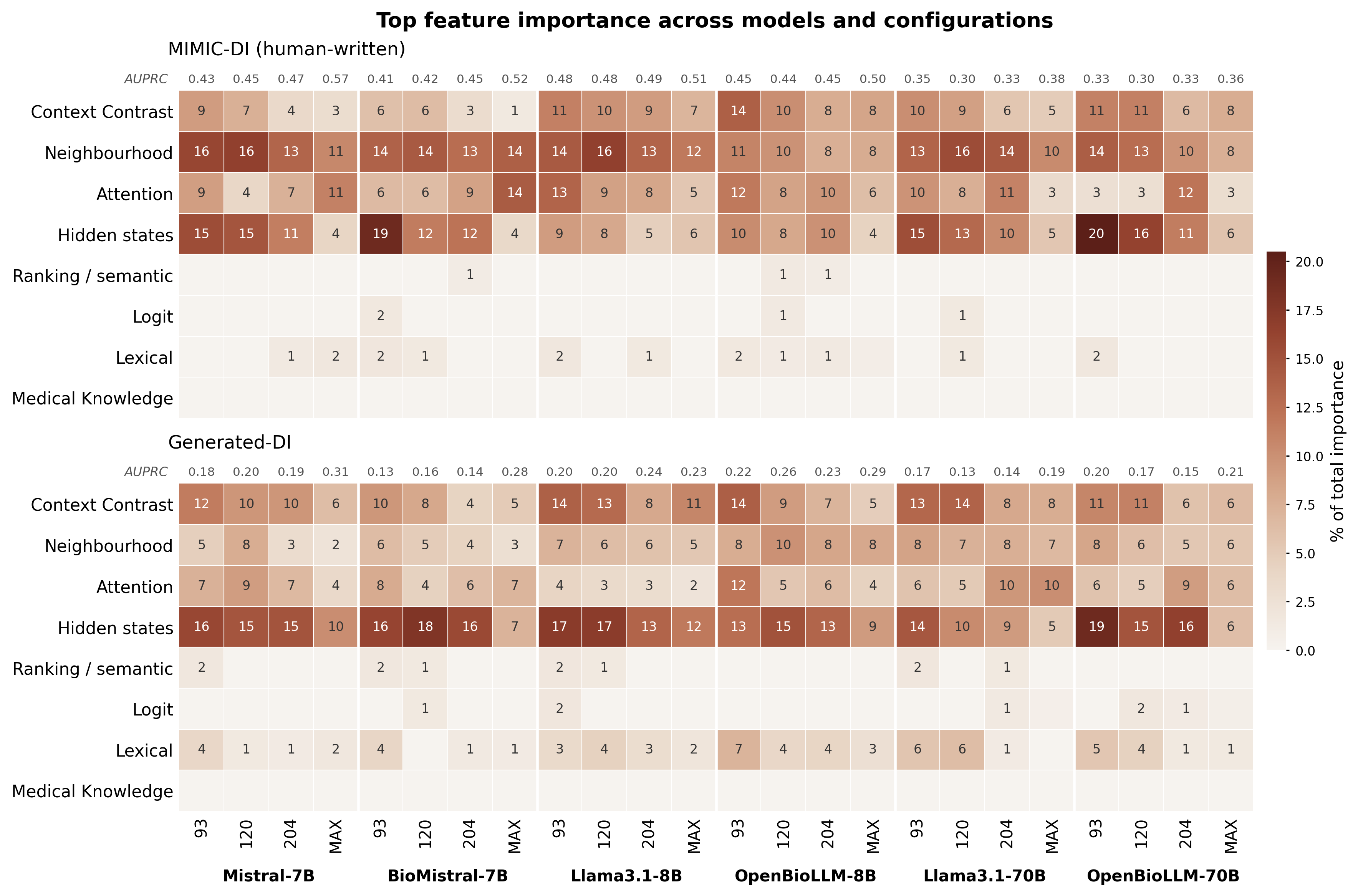}
    \caption{CatBoost Feature importance across models and feature configurations, with
importance measured by PredictionValuesChange. Cells give
the share of total importance carried by each family among the top 20
features. AUPRC
is shown above each column. MAX denotes 454 features for 7/8B models and 886
for 70B models.}
    \label{fig:catboost}
\end{figure*}

Figure~\ref{fig:catboost} repeats the analysis of
Figure~\ref{fig:heatmap} using CatBoost instead of XGBoost. The two
classifiers use different definitions of importance: XGBoost reports gain,
the average reduction in loss attributable to a feature across all splits,
while CatBoost reports PredictionValuesChange. 

The feature contribution is similar to the result from XGBoost. Distributional and internal
signals carry most of the weight, semantic features contribute little, and
medical NER features never enter the top 20 in any of the 48 configurations. \texttt{delta\_energy} ranks first again, with a mean share of 5.12\% against
2.67\% for \texttt{neighbor\_avg\_w5}. 

\section{Features Details}

\label{sec:features_details}

We extract token-level features from each model in four progressively richer configurations (93, 120, 204, and max: 454/886 features). To save space, each version is based on the previous one. All configurations share the same core groups: logit-based uncertainty signals, context-contrast features comparing distributions with and without the BHC, semantic similarity to top-$k$ predictions, neighborhood smoothness, medical entity detection, and lexical baseline statistics. Later versions expand these groups and gradually  introduce new ones. Tables~\ref{tab:features_93}--\ref{tab:features_454_886} record details of each configuration.

\begin{table*}[t]
\centering

\small
\begin{tabular}{p{2.1cm} p{3.7cm} p{6.5cm} c}
\toprule
\textbf{Group} & \textbf{Feature} & \textbf{Description / Formula} & \textbf{\#} \\
\midrule

% ---- Group 1 ----
\multirow{10}{2.1cm}{\centering Logit\\(with BHC)}
 & \texttt{entropy}             & $H = -\sum_v p_v \log p_v$ & \\
 & \texttt{normalized\_entropy} & $H / \log V$ & \\
 & \texttt{max\_prob}           & $\max_v p_v$ & \\
 & \texttt{current\_prob}       & $p_i$ & \\
 & \texttt{margin}              & $p_{(1)} - p_{(2)}$, top-2 probability gap & \\
 & \texttt{ratio}               & $p_{(1)} / p_{(2)}$ & \\
 & \texttt{topk\_cumulative}    & $\sum_{j=1}^{10} p_{(j)}$ & \\
 & \texttt{gini}                & $\frac{2\sum_j j\,p_{(j)}}{V \sum_j p_{(j)}} - \frac{V+1}{V}$,\; probabilities sorted descending & \\
 & \texttt{perplexity}          & $\exp(-\log p_i)$ & \\
 & \texttt{energy}              & $-\!\left(\max_v \ell_v + \log\sum_v e^{\,\ell_v - \max\ell}\right)$ & \multirow{-10}{*}{10} \\

\midrule

% ---- Group 2 ----
\multirow{3}{2.1cm}{\centering Context\\Contrast\\($\Delta$ BHC)}
 & \texttt{delta\_prob}    & $p_i - p^{-}_i$ & \\
 & \texttt{delta\_entropy} & $H(\mathbf{p}) - H(\mathbf{p}^{-})$ & \\
 & \texttt{delta\_energy}  & $\mathrm{energy}(\mathbf{p}) - \mathrm{energy}(\mathbf{p}^{-})$ & \multirow{-3}{*}{3} \\

\midrule

% ---- Group 3 ----
\multirow{7}{2.1cm}{\centering Ranking \&\\Semantic\\Similarity\\($k \in \{5,10,20\}$)}
 & \texttt{rank\_top$k$}           & Rank of token $i$ among top-$k$ predictions (0 if absent) & \\
 & \texttt{in\_top$k$}             & $\mathbf{1}[\mathrm{rank} > 0]$ & \\
 & \texttt{max\_sim\_top$k$}       & $\max_j \cos(\mathbf{e}_i,\, \mathbf{e}_{(j)})$ over top-$k$ tokens & \\
 & \texttt{avg\_sim\_top$k$}       & Mean cosine similarity to top-$k$ tokens & \\
 & \texttt{top3\_sim\_top$k$}      & Mean cosine similarity to top-3 tokens & \\
 & \texttt{sim\_std\_top$k$}       & Std.\ dev.\ of cosine similarities to top-$k$ & \\
 & \texttt{semantic\_rank\_top$k$} & First rank $j$ with $\cos(\mathbf{e}_i, \mathbf{e}_{(j)}) > 0.7$ & \multirow{-7}{*}{21} \\

\midrule

% ---- Group 4 ----
\multirow{5}{2.1cm}{\centering Neighbourhood\\($w \in \{2,3,5\}$)}
 & \texttt{neighbor\_avg\_w$w$}        & Mean $p_j$ of the $2w$ surrounding tokens & \\
 & \texttt{neighbor\_std\_w$w$}        & Std.\ dev.\ of surrounding token probabilities & \\
 & \texttt{isolation\_w$w$}            & $p_i - \bar{p}_\mathrm{nbr}$ & \\
 & \texttt{relative\_isolation\_w$w$}  & $(p_i - \bar{p}_\mathrm{nbr})\,/\,(\sigma_\mathrm{nbr} + \varepsilon)$ & \\
 & \texttt{medical\_density\_w$w$}     & Fraction of neighbors matching a medical keyword list & \multirow{-5}{*}{15} \\

\midrule

% ---- Group 5 ----
\parbox{2.1cm}{\centering Medical\\Keyword}
 & \texttt{is\_medical} & Binary flag: token string matches a curated medical keyword list & 1 \\

\midrule

% ---- Group 6 ----
\multirow{7}{2.1cm}{\centering Lexical \&\\Baseline}
 & \texttt{freq}              & Raw token frequency in the corpus & \\
 & \texttt{freq\_normalized}  & freq\,/\,total tokens & \\
 & \texttt{freq\_log}         & $\log(\mathrm{freq}+1)$ & \\
 & \texttt{idf}               & $\log(N\,/\,(df+1))$,\; $N$ = number of documents & \\
 & \texttt{rarity}            & $1\,/\,(\mathrm{freq}+1)$ & \\
 & \texttt{baseline\_prob}    & $p_i$ computed from a minimal BOS-only context & \\
 & \texttt{baseline\_entropy} & $H$ computed from a minimal BOS-only context & \multirow{-7}{*}{7} \\

\midrule

% ---- Group 7 ----
\multirow{5}{2.1cm}{\centering Hidden\\States\\(4 layers)}
 & \texttt{hidden\_norm\_l$\ell$} & $\|h_\ell\|_2$ & \\
 & \texttt{hidden\_mean\_l$\ell$} & $\mathrm{mean}(h_\ell)$ & \\
 & \texttt{hidden\_std\_l$\ell$}  & $\mathrm{std}(h_\ell)$ & \multirow{-3}{*}{12} \\
\cmidrule(l){2-4}
 & \multicolumn{2}{p{10cm}}{%
    \textit{4 layers sampled at uniform intervals.}
    8B: layers $\{8, 16, 24, 31\}$ (every 8 layers);
    70B: layers $\{20, 40, 60, 79\}$ (every 20 layers).} & \\

\midrule

% ---- Group 8 ----
\multirow{4}{2.1cm}{\centering Hidden\\State\\Change}
 & \texttt{layer\_change\_l$\ell_1$\_to\_l$\ell_2$} & $\|h_{\ell_2} - h_{\ell_1}\|_2$ between consecutive sampled layers & \multirow{-2}{*}{6} \\
 & \texttt{layer\_cosine\_l$\ell_1$\_to\_l$\ell_2$} & $\cos(h_{\ell_1},\, h_{\ell_2})$ & \\
\cmidrule(l){2-4}
 & \multicolumn{2}{p{10cm}}{%
    \textit{3 consecutive pairs from the 4 sampled layers above.}} & \\

\midrule

% ---- Group 9 ----
\multirow{5}{2.1cm}{\centering Attention\\Snapshots\\(6 heads)}
 & \texttt{attn\_entropy\_l$\ell$\_h$h$} & $H(\alpha_{\ell,h,i,\cdot})$, entropy of the attention row & \\
 & \texttt{attn\_to\_bhc\_l$\ell$\_h$h$} & $\sum_{j \leq L_\mathrm{BHC}} \alpha_{\ell,h,i,j}$, total attention weight to BHC tokens & \\
 & \texttt{attn\_max\_l$\ell$\_h$h$}     & $\max_j \alpha_{\ell,h,i,j}$ & \multirow{-3}{*}{18} \\
\cmidrule(l){2-4}
& \multicolumn{2}{p{10cm}}{%
    \textit{6 (layer, head) pairs sampled to cover shallow, mid, and deep
    layers.
    8B: $(7,16),(15,24),(23,0),(23,9),(23,28),(31,8)$;
    70B: $(7,16),(17,24),(57,0),(57,9),(57,28),(77,8)$.
    The sampling count is identical; indices are scaled to model depth and
    head count.}} & \\
\midrule
\multicolumn{3}{r}{\textbf{Total}} & \textbf{93} \\
\bottomrule
\end{tabular}

\caption{%
  93 token-level features.
  Parameters that differ
  between models are noted in the Description column.
  Notation: $p_i$ = softmax probability of the actual token $i$;
  $\mathbf{p}$ = full distribution over vocabulary $V$;
  $\ell_i$ = logit of token $i$;
  $\mathbf{p}^{-}$ = distribution computed \emph{without} BHC context;
  $h_\ell$ = hidden state vector at layer $\ell$;
  $\alpha_{\ell,h}$ = attention weights at layer $\ell$, head $h$.
}
\label{tab:features_93}
\end{table*}

\begin{table*}
\centering
\small
\begin{tabular}{p{2.1cm} p{3.5cm} p{6.5cm} c}
\toprule
\textbf{Group} & \textbf{Feature} & \textbf{Description / Formula} & \textbf{\#} \\
\midrule
Logit (with BHC) & \multicolumn{2}{l}{\textit{Same 10 features as Table~\ref{tab:features_93}.}} & 10\;/\;10 \\
\midrule
Context Contrast ($\Delta$ BHC) & \multicolumn{2}{l}{\textit{Same 3 features as Table~\ref{tab:features_93}.}} & 3\;/\;3 \\
\midrule
Ranking \& Semantic Similarity ($k\!\in\!\{5,10,20\}$) & \multicolumn{2}{l}{\textit{Same 21 features as Table~\ref{tab:features_93}.}} & 21\;/\;21 \\
\midrule
% ---- Neighbourhood: window set changed ----
\multirow{5}{2.1cm}{\centering Neighbourhood\\($w\!\in\!\{2,3,5,7\}$)}
 & \texttt{neighbor\_avg\_w$w$}       & Mean $p_j$ of the $2w$ surrounding tokens & \\
 & \texttt{neighbor\_std\_w$w$}       & Std.\ dev.\ of surrounding token probabilities & \\
 & \texttt{isolation\_w$w$}           & $p_i - \bar{p}_\mathrm{nbr}$ & \\
 & \texttt{relative\_isolation\_w$w$} & $(p_i - \bar{p}_\mathrm{nbr})\,/\,(\sigma_\mathrm{nbr}+\varepsilon)$ & \\
 & \texttt{medical\_density\_w$w$}    & Fraction of neighbours detected as medical terms (NER) & \multirow{-5}{*}{20\;/\;20} \\
\midrule
% ---- Medical NER: simplified vs 454/886 ----
\multirow{3}{2.1cm}{\centering Medical NER\\(scispaCy +\\UMLS/MedDRA)}
 & \texttt{is\_medical}         & Binary: token detected by NER or vocabulary & \\
 & \texttt{ner\_entity\_type}   & Integer-encoded entity type (12 classes: O, CHEMICAL, DISEASE, \ldots) & \\
 & \texttt{medical\_confidence} & Source confidence: NER+vocab\,=\,1.0,\; NER\,=\,0.85,\; vocab\,=\,0.6 & \multirow{-3}{*}{3\;/\;3} \\
\midrule
Lexical \& Baseline & \multicolumn{2}{l}{\textit{Same 7 features as Table~\ref{tab:features_93}.}} & 7\;/\;7 \\
\midrule
% ---- Hidden States: 8 sampled layers ----
\multirow{4}{2.1cm}{\centering Hidden\\States\\(8 layers)}
 & \texttt{hidden\_norm\_l$\ell$} & $\|h_\ell\|_2$ & \\
 & \texttt{hidden\_mean\_l$\ell$} & $\mathrm{mean}(h_\ell)$ & \\
 & \texttt{hidden\_std\_l$\ell$}  & $\mathrm{std}(h_\ell)$ & \multirow{-3}{*}{24\;/\;24} \\
\cmidrule(l){2-4}
 & \multicolumn{2}{p{10cm}}{\textit{8 layers sampled at uniform intervals (${\approx}$every 4 layers for 8B, every 10 for 70B). 8B: $\{2,6,10,14,18,22,26,31\}$;\enspace 70B: $\{5,15,25,35,45,55,65,79\}$.}} & \\
\midrule
% ---- Hidden State Change: 7 pairs ----
\multirow{3}{2.1cm}{\centering Hidden\\State\\Change\\(7 pairs)}
 & \texttt{layer\_change\_l$\ell_1$\_to\_l$\ell_2$} & $\|h_{\ell_2} - h_{\ell_1}\|_2$ between consecutive sampled layers & \multirow{-2}{*}{14\;/\;14} \\
 & \texttt{layer\_cosine\_l$\ell_1$\_to\_l$\ell_2$} & $\cos(h_{\ell_1},\, h_{\ell_2})$ & \\
\cmidrule(l){2-4}
 & \multicolumn{2}{p{10cm}}{\textit{7 consecutive pairs from the 8 sampled layers above.}} & \\
\midrule
% ---- Attention Snapshots: 6 heads ----
\multirow{4}{2.1cm}{\centering Attention\\Snapshots\\(6 heads)}
 & \texttt{attn\_entropy\_l$\ell$\_h$h$} & $H(\alpha_{\ell,h,i,\cdot})$, entropy of the attention row & \\
 & \texttt{attn\_to\_bhc\_l$\ell$\_h$h$} & $\sum_{j \leq L_\mathrm{BHC}} \alpha_{\ell,h,i,j}$, total attention weight to BHC & \\
 & \texttt{attn\_max\_l$\ell$\_h$h$}     & $\max_j \alpha_{\ell,h,i,j}$ & \multirow{-3}{*}{18\;/\;18} \\
\cmidrule(l){2-4}
 & \multicolumn{2}{p{10cm}}{\textit{6 (layer, head) pairs covering shallow, mid, and deep layers. 8B: $(7,16),(15,24),(23,0),(23,9),(23,28),(31,8)$;\enspace 70B: $(37,24),(17,16),(57,0),(57,9),(57,28),(77,8)$. Layer indices scaled proportionally; head count identical.}} & \\
\midrule
\multicolumn{3}{r}{\textbf{Total}} & \textbf{120\;/\;120} \\
\bottomrule
\end{tabular}
\caption{%
  Token-level features (120 total) extracted using
  \textbf{Llama-3.1-8B-Instruct} (32 layers, 32 attention heads) and
  \textbf{Llama3-OpenBioLLM-70B} (80 layers, 64 attention heads).
  Groups identical to Table~\ref{tab:features_93} are abbreviated;
  only modified or new groups are shown in full.
  Compared to the 93-feature set: (1) Increased windown size 7; (2) medical detection upgraded from
  keyword matching to scispaCy NER with 3 output features; (3) hidden-state sampling expanded to 8 layers
  with 7 inter-layer change pairs.
  Notation as in Table~\ref{tab:features_93};
  layer/head indices that differ between models are listed in the notes rows.
}
\label{tab:features_120}
\end{table*}

\begin{table*}
\centering
\small
\begin{tabular}{p{2.3cm} p{3.2cm} p{7.2cm} c}
\toprule
\textbf{Group} & \textbf{Added / Changed} & \textbf{Description / Formula} & \textbf{120\;/\;204} \\
\midrule
\multicolumn{3}{p{12.7cm}}{\textit{Unchanged from Table~\ref{tab:features_120}: logit (10), ranking \& semantic (21), neighbourhood (20), lexical \& baseline (7), hidden states (24), hidden state change (14), attention snapshots (18).}} & 114\;/\;114 \\
\midrule
Context Contrast & \texttt{kl\_with\_vs\_no}, \texttt{kl\_no\_vs\_with}, \texttt{js\_divergence}, \texttt{rank\_with\_bhc}, \texttt{rank\_without\_bhc}, \texttt{delta\_rank}, \texttt{topk\_overlap}, \texttt{topk\_jaccard}, \texttt{delta\_margin}, \texttt{prob\_vs\_no\_max} & Divergence, rank shift, and top-$k$ overlap between $\mathbf{p}$ and $\mathbf{p}^-$, complementing the three $\Delta$ features carried over from Table~\ref{tab:features_120} & 3\;/\;13 \\
\midrule
Medical NER & 6 one-hot entity-type flags, \texttt{ner\_is\_high\_risk} & Adds entity-type-specific flags (chemical, disease, gene, cancer, anatomy, pathology) and a high-risk composite flag on top of the \texttt{is\_medical} / \texttt{ner\_entity\_type} / \texttt{medical\_confidence} features already present in Table~\ref{tab:features_120} & 3\;/\;10 \\
\midrule
PMI \textit{(new)} & \texttt{pmi}, \texttt{pmi\_vs\_prior}, \texttt{npmi}, \texttt{cpmi} & $\log p^\mathrm{with}_i-\log p^\mathrm{no}_i$; variants against the language prior, length-normalised, and standardised within entity type & 0\;/\;4 \\
\midrule
Entropy Decomposition \textit{(new)} & \texttt{bhc\_info\_gain}, \texttt{ctx\_info\_gain}, \texttt{total\_info\_gain}, \texttt{bhc\_contribution\_ratio}, \texttt{bhc\_info\_gain\_norm} & Separates the entropy reduction attributable to the BHC from that of the surrounding context, using $H(\mathbf{p}^\mathrm{prior})$, $H(\mathbf{p}^-)$, $H(\mathbf{p})$ & 0\;/\;5 \\
\midrule
Medical Prior Decomposition \textit{(new)} & \texttt{halluc\_risk\_ratio}, \texttt{patient\_specificity}, \texttt{context\_reliance\_score}, \texttt{prob\_dominance\_order}, \texttt{patient\_specificity\_norm} & Distinguishes tokens supported by patient-specific evidence from those recoverable from medical prior alone & 0\;/\;5 \\
\midrule
\multirow{3}{2.3cm}{Attention Drift \textit{(new)}}
 & \texttt{attn\_drift\_l$\ell_1$\_l$\ell_2$\_h$h$} & $D_\mathrm{KL}(\tilde\alpha_{\ell_1,h,i}\,\|\,\tilde\alpha_{\ell_2,h,i})$ between adjacent-layer attention rows; captures sudden shifts in grounding & \multirow{-3}{*}{0\;/\;44} \\
\cmidrule(l){2-4}
 & \multicolumn{2}{p{10.4cm}}{\textit{Three bands of 4 consecutive layers
 (shallow, mid, deep) give 12 layers and 11 adjacent pairs;
 $11\times4$ heads $=44$.
 8B: $\{0\text{--}3\}\cup\{14\text{--}17\}\cup\{28\text{--}31\}$,
 $h\!\in\!\{0,8,16,24\}$.
 70B: $\{0\text{--}3\}\cup\{35\text{--}38\}\cup\{76\text{--}79\}$,
 $h\!\in\!\{0,16,32,48\}$.}} & \\
\midrule
\multirow{3}{2.3cm}{Attention Rollout \textit{(new)}}
 & \texttt{rollout\_to\_bhc\_l$\ell$}, \texttt{rollout\_entropy\_l$\ell$}, \texttt{rollout\_max\_weight\_l$\ell$} & $\mathbf{R}_\ell=\prod_{k=0}^{\ell}(0.5\,\mathbf{A}_k+0.5\,\mathbf{I})$, row-normalised at each step; cumulative attention to BHC, its entropy, and its max weight & \multirow{-3}{*}{0\;/\;9} \\
\cmidrule(l){2-4}
 & \multicolumn{2}{p{10.4cm}}{\textit{One checkpoint at the last layer of each
 band (3 checkpoints $\times$ 3 features). 8B: $\ell\!\in\!\{3,17,31\}$;
 70B: $\ell\!\in\!\{3,38,79\}$.}} & \\
\midrule
\multicolumn{3}{r}{\textbf{Total}} & \textbf{120\;/\;204} \\
\bottomrule
\end{tabular}
\caption{%
  The 204-feature set changed from the 120-feature set of Table~\ref{tab:features_120}.
  Groups marked \textit{(new)} did not exist in the 120-feature set (hence the $0$ in the left-hand count) and were introduced for the first time here.
  $\mathbf{p}$, $\mathbf{p}^-$, and $\mathbf{p}^\mathrm{prior}$ denote the output distribution with the BHC, without the BHC, and from a minimal BOS-only context.
  Note that \texttt{delta\_entropy} and \texttt{bhc\_info\_gain} are sign-inverted forms of the same quantity.
}
\label{tab:features_204}
\end{table*}

\begin{table*}
\centering
\small
\begin{tabular}{p{3.0cm} p{3.3cm} p{6.0cm} c}
\toprule
\textbf{Group} & \textbf{Feature} & \textbf{Description / Formula} & \textbf{\#} \\
\midrule
Logit / Contrast / Ranking / Neighbourhood / Medical NER / Lexical / PMI / Entropy Decomp.\ / Medical Prior Decomp. & \multicolumn{2}{l}{\textit{Same as Table~\ref{tab:features_204}: 54 + 10 + 7 + 3 + 5 + 5 = 84 features.}} & 84\;/\;84 \\
\midrule
% ---- Hidden States: all L layers ----
\multirow{4}{2.6cm}{\centering Hidden\\States\\(all $L$ layers,\\expanded)}
 & \texttt{hidden\_norm\_l$\ell$} & $\|h_\ell\|_2$,\enspace $\ell=0,\ldots,L{-}1$ & \\
 & \texttt{hidden\_mean\_l$\ell$} & $\mathrm{mean}(h_\ell)$ & \\
 & \texttt{hidden\_std\_l$\ell$}  & $\mathrm{std}(h_\ell)$ & \multirow{-3}{*}{96\;/\;240} \\
\cmidrule(l){2-4}
 & \multicolumn{2}{p{9.3cm}}{\textit{Expanded from 8 sampled layers to all $L$ layers. 8B: $32\times3=96$;\enspace 70B: $80\times3=240$.}} & \\
\midrule
% ---- Hidden State Change: all L-1 pairs ----
\multirow{3}{2.6cm}{\centering Hidden\\State Change\\(all $L{-}1$ pairs,\\expanded)}
 & \texttt{layer\_change\_l$\ell$\_to\_l$(\ell{+}1)$} & $\|h_{\ell+1}-h_\ell\|_2$ & \multirow{-2}{*}{62\;/\;158} \\
 & \texttt{layer\_cosine\_l$\ell$\_to\_l$(\ell{+}1)$} & $\cos(h_\ell,\,h_{\ell+1})$ & \\
\cmidrule(l){2-4}
 & \multicolumn{2}{p{9.3cm}}{\textit{Expanded from 7 sampled pairs to all consecutive pairs. 8B: $31\times2=62$;\enspace 70B: $79\times2=158$.}} & \\
\midrule
% ---- Attention Snapshots: unchanged ----
Attention Snapshots (6 heads) & \multicolumn{2}{l}{\textit{Same 18 features as Table~\ref{tab:features_204}.}} & 18\;/\;18 \\
\midrule
% ---- Attention Drift: all adjacent pairs ----
\multirow{4}{2.6cm}{\centering Attention\\Drift\\(4 heads,\\all $L{-}1$ pairs,\\expanded)}
 & \texttt{attn\_drift\_l$\ell$\_l$(\ell{+}1)$\_h$h$} &
   $D_\mathrm{KL}(\tilde\alpha_{\ell,h,i}\,\|\,\tilde\alpha_{\ell+1,h,i})$,
   KL divergence between adjacent-layer attention rows & \multirow{-4}{*}{124\;/\;316} \\
\cmidrule(l){2-4}
 & \multicolumn{2}{p{9.3cm}}{\textit{Expanded from 3 bands to all adjacent layer pairs. 4 heads at uniform intervals. 8B: $h\!\in\!\{0,8,16,24\}$, $31$ pairs $=124$;\enspace 70B: $h\!\in\!\{0,16,32,48\}$, $79$ pairs $=316$.}} & \\
\midrule
% ---- Rollout: 8 checkpoints ----
\multirow{4}{2.6cm}{\centering Attention\\Rollout\\(8 checkpoints,\\expanded)}
 & \texttt{rollout\_to\_bhc\_l$\ell$}     & $\sum_{j<L_\mathrm{BHC}}[\mathbf{R}_\ell]_{i,j}$, cumulative attention to BHC & \\
 & \texttt{rollout\_entropy\_l$\ell$}     & $H(\mathbf{R}_\ell[i,\cdot])$ & \\
 & \texttt{rollout\_max\_weight\_l$\ell$} & $\max_j[\mathbf{R}_\ell]_{i,j}$ & \multirow{-3}{*}{24\;/\;24} \\
\cmidrule(l){2-4}
 & \multicolumn{2}{p{9.3cm}}{\textit{$\mathbf{R}_\ell=\prod_{k=0}^{\ell}(0.5\,\mathbf{A}_k+0.5\,\mathbf{I})$; expanded from 3 to 8 uniformly spaced checkpoints (8B: every 4 layers; 70B: every 10 layers).}} & \\
\midrule
\multicolumn{3}{r}{\textbf{Total}} & \textbf{454\;/\;886} \\
\bottomrule
\end{tabular}
\caption{%
  Token-level features extracted using
  \textbf{Llama-3.1-8B-Instruct} (\textbf{454 features}, 32 layers, 32 attention heads) and
  \textbf{Llama3-OpenBioLLM-70B} (\textbf{886 features}, 80 layers, 64 attention heads).
  Groups unchanged from Table~\ref{tab:features_204} are abbreviated.
  Compared to the 204-feature set, three groups are expanded to cover all model layers:
  (1) hidden-state features extended from 8 sampled layers to all $L$ layers;
  (2) hidden-state change features extended from 7 sampled pairs to all $L{-}1$ consecutive pairs;
  (3) attention drift extended from 3 depth bands to all adjacent layer pairs;
  (4) rollout checkpoints increased from 3 to 8.
  The increase in total features (204$\to$454/886) is entirely due to these layer-coverage expansions.
  Notation as in Table~\ref{tab:features_93}.
}
\label{tab:features_454_886}
\end{table*}

\section{Full Experimental Results}

\label{sec:complete_results}
This appendix reports complete results for all experiments. Table~\ref{tab:appendix_full_results} presents Reverse Probing results for all six base LLMs and all configurations. 

\begin{table*}[p]
\small
\centering
\begin{tabular}{lrrr@{\hspace{3em}}rrr}
\toprule
& \multicolumn{3}{c}{\textbf{XGBoost}} & \multicolumn{3}{c}{\textbf{CatBoost}} \\
\cmidrule(lr){2-4} \cmidrule(lr){5-7}
\textbf{Config} & \textbf{F1} & \textbf{AUCROC} & \textbf{AUPRC} & \textbf{F1} & \textbf{AUCROC} & \textbf{AUPRC} \\
\midrule
\multicolumn{7}{l}{\textbf{OpenBioLLM-8B / MIMIC-DI}} \\
93            & 0.3953 & 0.8598 & 0.4418 & 0.4015 & 0.8645 & 0.4485 \\
120           & 0.4466 & 0.8781 & 0.4615 & 0.4259 & 0.8643 & 0.4445 \\
204           & 0.4454 & 0.8774 & 0.4600 & 0.4247 & 0.8684 & 0.4525 \\
MAX (454)     & 0.4651 & 0.8833 & 0.4976 & 0.4534 & 0.8848 & 0.5029 \\
\multicolumn{7}{l}{\textbf{OpenBioLLM-8B / Generated-DI}} \\
93            & 0.2162 & 0.8714 & 0.1719 & 0.2273 & 0.8818 & 0.2194 \\
120           & 0.2857 & 0.9046 & 0.2421 & 0.4082 & 0.8739 & 0.2588 \\
204           & 0.2449 & 0.8983 & 0.1897 & 0.3396 & 0.8773 & 0.2277 \\
MAX (454)     & 0.3038 & 0.9364 & 0.2892 & 0.3146 & 0.9144 & 0.2895 \\
\midrule
\multicolumn{7}{l}{\textbf{Llama3.1-8B / MIMIC-DI}} \\
93            & 0.4608 & 0.8657 & 0.4742 & 0.4435 & 0.8527 & 0.4842 \\
120           & 0.4272 & 0.8814 & 0.4732 & 0.4504 & 0.8659 & 0.4808 \\
204           & 0.4488 & 0.8747 & 0.4763 & 0.4331 & 0.8724 & 0.4858 \\
MAX (454)     & 0.4712 & 0.8967 & 0.5110 & 0.4766 & 0.8755 & 0.5140 \\
\multicolumn{7}{l}{\textbf{Llama3.1-8B / Generated-DI}} \\
93            & 0.2276 & 0.9133 & 0.1496 & 0.2143 & 0.8481 & 0.1956 \\
120           & 0.2000 & 0.9228 & 0.1854 & 0.2609 & 0.9136 & 0.2033 \\
204           & 0.2222 & 0.9229 & 0.1479 & 0.2600 & 0.9340 & 0.2761 \\
MAX (454)     & 0.3117 & 0.9365 & 0.2121 & 0.3036 & 0.9245 & 0.2253 \\
\midrule
\multicolumn{7}{l}{\textbf{OpenBioLLM-70B / MIMIC-DI}} \\
93            & 0.3665 & 0.8545 & 0.3023 & 0.3900 & 0.8658 & 0.3260 \\
120           & 0.3672 & 0.8512 & 0.2991 & 0.3737 & 0.8562 & 0.3023 \\
204           & 0.3890 & 0.8639 & 0.3125 & 0.3870 & 0.8585 & 0.3281 \\
MAX (886)     & 0.3876 & 0.8710 & 0.3500 & 0.4009 & 0.8706 & 0.3597 \\
\multicolumn{7}{l}{\textbf{OpenBioLLM-70B / Generated-DI}} \\
93            & 0.2222 & 0.8089 & 0.1249 & 0.2540 & 0.8262 & 0.2008 \\
120           & 0.1636 & 0.8130 & 0.1208 & 0.2286 & 0.8218 & 0.1725 \\
204           & 0.1837 & 0.8141 & 0.1277 & 0.2162 & 0.7998 & 0.1548 \\
MAX (886)     & 0.3030 & 0.7985 & 0.2162 & 0.2459 & 0.8117 & 0.2073 \\
\midrule
\multicolumn{7}{l}{\textbf{Llama3.1-70B / MIMIC-DI}} \\
93            & 0.3657 & 0.8407 & 0.2965 & 0.3839 & 0.8494 & 0.3528 \\
120           & 0.3382 & 0.8408 & 0.3057 & 0.3502 & 0.8337 & 0.2964 \\
204           & 0.3643 & 0.8524 & 0.3233 & 0.3636 & 0.8533 & 0.3268 \\
MAX (886)     & 0.3932 & 0.8594 & 0.3934 & 0.3968 & 0.8563 & 0.3835 \\
\multicolumn{7}{l}{\textbf{Llama3.1-70B / Generated-DI}} \\
93            & 0.1667 & 0.8344 & 0.1252 & 0.2121 & 0.8055 & 0.1679 \\
120           & 0.2609 & 0.8226 & 0.1486 & 0.1875 & 0.7657 & 0.1322 \\
204           & 0.1633 & 0.7877 & 0.1247 & 0.1905 & 0.8179 & 0.1378 \\
MAX (886)     & 0.2963 & 0.8132 & 0.1697 & 0.2326 & 0.7999 & 0.1939 \\
\midrule
\multicolumn{7}{l}{\textbf{BioMistral-7B / MIMIC-DI}} \\
93            & 0.3882 & 0.8293 & 0.4297 & 0.3980 & 0.8216 & 0.4126 \\
120           & 0.3834 & 0.8482 & 0.4366 & 0.3560 & 0.8328 & 0.4197 \\
204           & 0.4057 & 0.8615 & 0.4562 & 0.4110 & 0.8512 & 0.4543 \\
MAX (454)     & 0.4812 & 0.8911 & 0.5478 & 0.4535 & 0.8691 & 0.5178 \\
\multicolumn{7}{l}{\textbf{BioMistral-7B / Generated-DI}} \\
93            & 0.2020 & 0.8192 & 0.1170 & 0.1765 & 0.7856 & 0.1254 \\
120           & 0.2222 & 0.8364 & 0.1997 & 0.2254 & 0.7979 & 0.1623 \\
204           & 0.2785 & 0.8318 & 0.1833 & 0.2432 & 0.8062 & 0.1442 \\
MAX (454)     & 0.3103 & 0.8819 & 0.2316 & 0.2712 & 0.8435 & 0.2805 \\
\midrule
\multicolumn{7}{l}{\textbf{Mistral-7B / MIMIC-DI}} \\
93            & 0.4459 & 0.8480 & 0.4654 & 0.3944 & 0.8260 & 0.4328 \\
120           & 0.4514 & 0.8598 & 0.4818 & 0.4378 & 0.8372 & 0.4545 \\
204           & 0.4742 & 0.8594 & 0.4894 & 0.4259 & 0.8493 & 0.4706 \\
MAX (454)     & 0.5519 & 0.9023 & 0.6022 & 0.4928 & 0.8937 & 0.5737 \\
\multicolumn{7}{l}{\textbf{Mistral-7B / Generated-DI}} \\
93            & 0.2500 & 0.8069 & 0.1938 & 0.2740 & 0.8178 & 0.1799 \\
120           & 0.3000 & 0.8162 & 0.1878 & 0.2963 & 0.8397 & 0.1960 \\
204           & 0.3125 & 0.8164 & 0.2005 & 0.3291 & 0.8030 & 0.1932 \\
MAX (454)     & 0.3611 & 0.8476 & 0.2373 & 0.3934 & 0.8539 & 0.3070 \\
\bottomrule
\end{tabular}
\caption{Reverse Probing full results across all models.}
\label{tab:appendix_full_results}
\end{table*}

\end{document}